%% file: main.tex
\documentclass[letterpaper, 10 pt, conference]{ieeeconf}  

\IEEEoverridecommandlockouts                              

\usepackage{cite}
\usepackage{subcaption}
\usepackage{url}
\usepackage{hyperref}%
\usepackage{graphics}
\usepackage{amsmath}
\usepackage{amssymb}
\usepackage{nameref}
\usepackage{color}
\usepackage{comment}
\usepackage{mathtools}

\input wenmacro.tex
\input robmacro.tex

\title{
Industrial Robot Trajectory Tracking Using Multi-Layer Neural Networks Trained by Iterative Learning Control
}

\author{Shuyang Chen$^{1}$ and John T. Wen$^{2}$, \textit{Fellow, IEEE}
\thanks{This research was sponsored in part by the Office of the Secretary of Defense
under Agreement Number W911NF-17-3-0004 
and in part by the New York State Matching Grant program and by the Center for Automation Technologies and Systems (CATS)  at Rensselaer Polytechnic Institute under a block grant from the New York State Empire State Development Division of Science, Technology and
Innovation (NYSTAR).}
\thanks{$^{1}$Shuyang Chen is with the Department of Mechanical Engineering, Rensselaer Polytechnic Institute, 110 8th St, Troy, NY 12180, USA
        {\tt\small chens26@rpi.edu}}%
\thanks{$^{2}$John T. Wen is with the Department
of Electrical, Computer \& Systems Engineering, Rensselaer Polytechnic Institute, Troy, NY 12180 USA.
        {\tt\small wenj@rpi.edu}}%
}

\begin{document}

\maketitle
\thispagestyle{empty}
\pagestyle{empty}

\begin{abstract}

Fast and precise robot motion is needed in certain applications such as electronic manufacturing, additive manufacturing and assembly. 
Most industrial robot motion controllers allow externally commanded motion profile, but the trajectory tracking performance is affected by the robot dynamics and joint servo controllers which users have no direct access and little information.  The performance is further compromised by time delays in transmitting the external command as a setpoint to the inner control loop.
This paper presents an approach of combining neural networks and iterative learning control to improve the trajectory tracking performance for a multi-axis articulated industrial robot.
For a given desired trajectory, the external command is iteratively refined using a high fidelity dynamical simulator to compensate for the robot inner loop dynamics. 
These desired trajectories and the corresponding refined input trajectories are then used to train multi-layer neural networks to emulate the dynamical inverse of the nonlinear inner loop dynamics. 
%
We show that with a sufficiently rich training set, the trained neural networks can generalize well to trajectories beyond the training set. 
In applying the trained neural networks to the physical robot, the tracking performance still improves but not as much as in the simulator.  
We show that transfer learning can effectively bridge the gap between simulation and the physical robot. In the end, we test the trained neural networks on other robot models in simulation and demonstrate the possibility of a general purpose network. Development and evaluation of this methodology is based on the ABB IRB6640-180 industrial robot and ABB RobotStudio software packages.
%
\end{abstract}

\begin{keywords}
Deep learning in robotics and automation, industrial robots, motion control, iterative learning control
\end{keywords}

\section{INTRODUCTION}

Robots have been widely utilized in industrial tasks including assembly, welding, painting, packaging and labeling~\cite{review2008}. In many cases they are controlled to track a given trajectory by an external motion command interface, which is available for many industrial robot controllers, including 
High Speed Controller (HSC) of Yasakawa Motoman (2~ms) \cite{MotomanController}, Low Level Interface (LLI) for St\"aubli (4~ms) \cite{StaubliController}, Robot Sensor Interface (RSI) of Kuka (12~ms) \cite{KukaController}, and Externally Guided Motion (EGM) of ABB (4~ms)~\cite{irc5_software_manual}. Although high-precision industrial robots have been well established in manufacturing and fabrication that require precise motion control such as aerospace assembly,
laser scanning and operation in crowded and unstructured environment for decades, it still remains a challenge to track a given trajectory fast and accurately. In general, a trade-off exists between reduction of cycle time and improvement of tracking accuracy for industrial robots.


\begin{figure}[tb]
\centering
\includegraphics[width=0.45\textwidth]{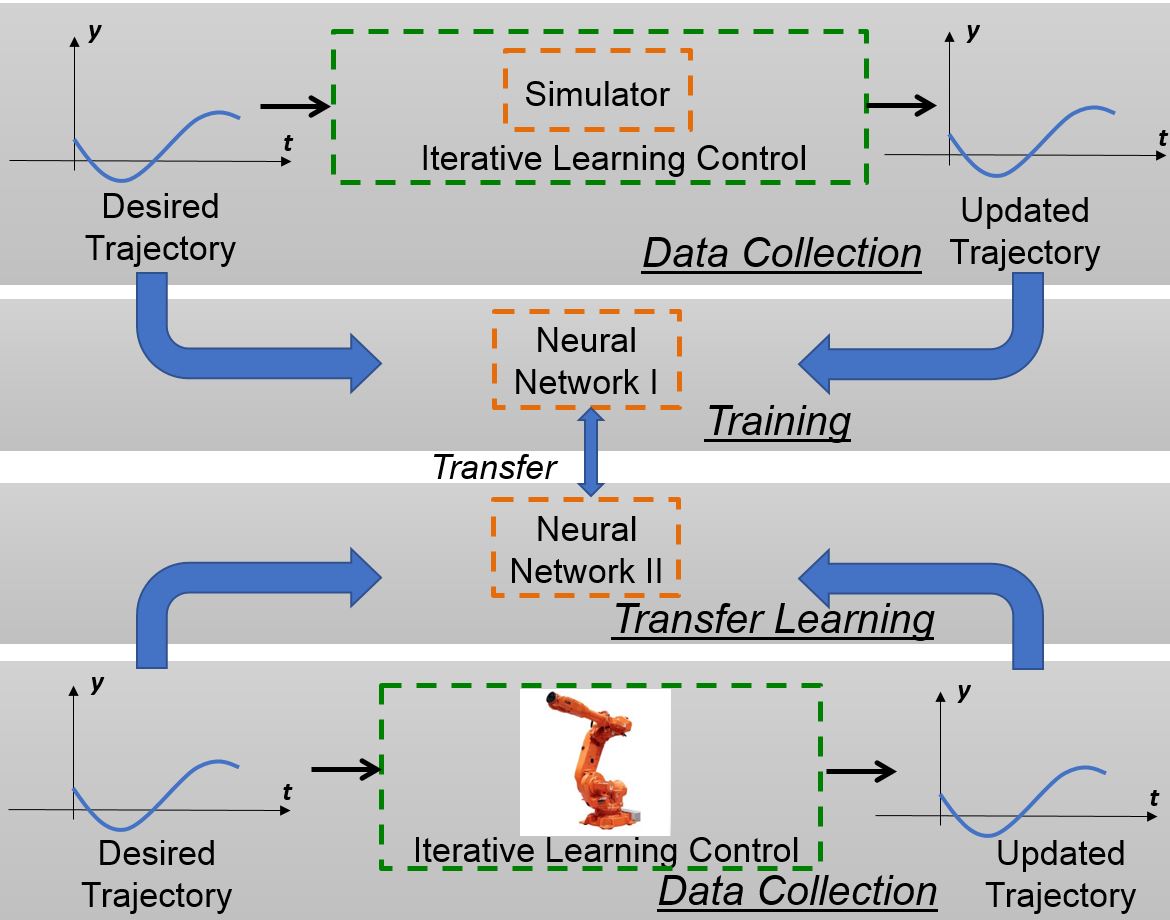}
\caption{Structure of the proposed trajectory tracking approach.}
\label{fig:entire_framework} 
\end{figure}

Trajectory tracking control for nonlinear systems via dynamics compensation is a well-studied problem. In order to compensate for the nonlinear system dynamics, most of the cases the simplified dynamic models are obtained with some assumptions, and then either feedback control~\cite{Marino1993}, backstepping control~\cite{Frazzoli2000, Otani2006}, sliding mode control~\cite{Kim1998}, model predictive control~\cite{Conceicao2010}, feedback linearization~\cite{kayacan_bayraktaroglu_saeys_2012}, variable structure control (VSC)~\cite{Sankaranarayanan1997}, differential dynamic programming (DDP)~\cite{Tassa2014} and iterative learning control (ILC)~\cite{Moore1999, Potsaid2007} will be applied to the simplified dynamic models, or the dynamic inversion is derived~\cite{Bayo1989, MacKunis2010, Li2014}. However, these approaches are not feasible in practice on highly complex nonlinear systems with unmodeled dynamics and uncertainties.

To address this problem, approximation-based control strategies have been developed based on fuzzy logic system~\cite{Cheng2012} and neural networks (NNs). NNs, with their expressive power of approximation widely recognized, have been utilized to represent direct dynamics, inverse dynamics, or a certain nonlinear mapping~\cite{Corradini2003, Velagic2012NeuralNC, WEI2015520}. In general, NNs-based dynamics compensation can be classified into two main categories: either NN is used to produce control inputs for a feedback controller that is designed based on a nominal robot dynamic model, or NN directly estimates the dynamic system inversion. A detailed review of NNs-based dynamics compensation is discussed in Section~\ref{Related Work} and the contributions of this paper is also highlighted.


The goal of this work is to design a feedforward controller to compensate for the inner loop dynamics without on-line iterations so that an industrial robot can accurately track specified trajectories.
Our approach, as illustrated in Fig.~\ref{fig:entire_framework}, combines iterative learning control (ILC) with multi-layer neural networks.  
Specifically, we implement ILC on a large number of trajectories over the robot workspace to obtain input trajectories that minimize the tracking errors.
This set of desired/input trajectories is used to train a neural network off-line (Neural Network~I, {\em NN-I}, in Fig.~\ref{fig:entire_framework}) to enable generalization beyond the training set.
For safety and speed, the ILC is performed in a dynamical simulator of the robot.  However, when the NN feedforward control is applied to the physical robot, the tracking performance degrades due to the invariable discrepancy between simulation and reality. To narrow this reality gap, we fine-tune the {\em NN-I} by collecting data of implementing ILC on the physical robot for a moderate amount of trajectories, inspired by the idea of transfer learning\cite{Pan2010}. The updated  NN (Neural Network~II, {\em NN-II}, in Fig.~\ref{fig:entire_framework}) shows improved tracking performance on the physical robot. Last but not the least, we apply the trained NNs ({\em NN-I}) to other two robot models in simulation and both of them show much improved tracking performance. Simulation and experimental results using the ABB RobotStudio software and ABB IRB6640-180 robot are provided to demonstrate the feasibility of our approach.

The rest of the paper is organized as follows. Section~\ref{Related Work} summarizes
related work on NNs-based dynamics compensation for trajectory tracking control and our contributions. Section~\ref{problem statement} states the problem, followed by the methodology in Section~\ref{methodology}. The simulation and experimental results are presented in Section~\ref{results}. Finally, Section~\ref{conclusions and future work} concludes the paper and adds some new insight.

\section{Related Work}
\label{Related Work}

In recent years, controller design based on NNs has attracted lots of attention due to their strong generalization capability with the availability of large amount of training data and powerful computational device~\cite{Kober2013}. In summary, the NN-based controller design mostly falls within the context of reinforcement learning (RL), predictive control, optimal control and adaptive control. And the NN is used either to approximate policy functions~\cite{Endo2008, Geng2006, Zhang2016} and value functions~\cite{duan2008, Riedmiller2009}, mimic a control law~\cite{WAI2003425}, or approximate the dynamic systems.

For tasks of trajectory tracking, model-based NNs controllers design have been proposed with NNs generating control inputs
for a torque-based robot controller that is designed based on a nominal robot dynamics model approximated by NNs. In~\cite{Lewis1996, Fierro1998, Oliveira2000}, a torque level feedback controller was developed, where the multilayer feedforward NNs were used to approximate the dynamics of the robot. Radial basis function (RBF) NN for system approximation is another popular choice in adaptive controller design~\cite{Lin2001, Wang2012, Hua2013, Dai2014, Yang2014, Wang2015, Han2016, He2016, Yang2017}. In~\cite{Yang2017}, an adaptive control strategy in joint torque level using RBF NNs for dynamics approximation was proposed to compensate for the unknown robot dynamics and heavy payload. Similar work using RBF NNs for feedback controllers design to minimize the trajectory tracking error between the master and the slave robots was also reported in~\cite{Hua2013}. Typically, it requires to train a large amount of parameters to approximate the robot dynamic model using the RBF NNs. Sun et al.~\cite{SUN20112377} and Park et al.~\cite{Park2009} proposed a NN-based sliding mode adaptive controller that combines sliding mode control and NNs approximation for trajectory tracking. In~\cite{He2018}, a fuzzy NN control algorithm was developed using impedance learning for a constrained robot. Other choices of NNs include fuzzy wavelet NN~\cite{Lu2011}, and recurrent NN~\cite{
Perez-Cruz2014}. In the above works, NN-based feedback controller was proposed and stability analysis such as Lyapunov stability theory was applied to the closed-form system to guarantee the bounded tracking error and derive the training law of the NNs. Thus, the NNs parameters were updated on-line and it took some time before the tracking error converged.  Although the result is impressive, the designed control architecture is very complicated and computationally expensive.

\begin{figure}[tb]
\centering
\includegraphics[width=0.42\textwidth]{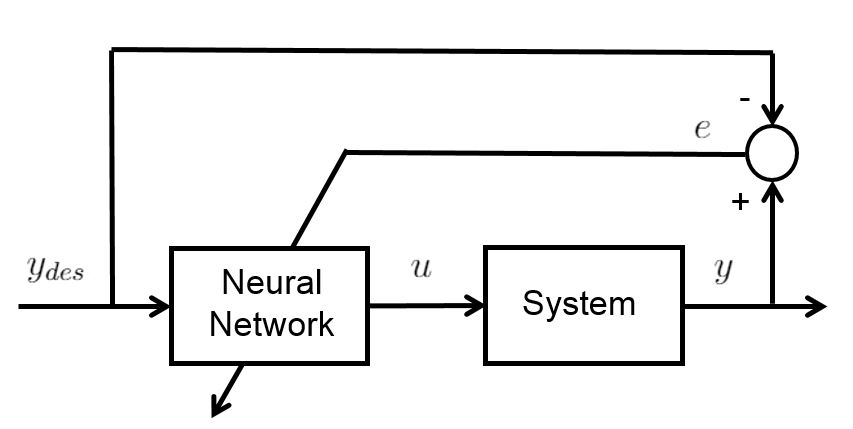}
\caption{General NN controller diagram with NN as the system inverse.}
\label{fig:nn_flow} 
\end{figure}

Works have also been reported of using NNs as an approximation of the inverse of the real plant (either or not under a feedback loop). Fig.~\ref{fig:nn_flow} shows a typical architecture for a direct inverse NN controller, with the output of NN $u$ driving the output of the system to the desired one $y_{des}$. Here, the NN acts effectively as the inverse of the plant and is trained by the error between the desired and actual responses of the system using gradient descent. However, the update of the NNs parameters requires the Jacobian of the plant so either a plant dynamic model~\cite{Abiyev2008}, or an estimation of the derivative of the error with respect to the plant~\cite{Psaltis1988} must be obtained, which is usually unavailable in practice. To avoid the estimation of the system Jacobian, a direct inverse NN was coupled with a feedback controller and the NN was trained by minimizing the feedback signal so that the estimation of system Jacobian can be avoid~\cite{Miyamoto1988}. Then, NN alone acts as the dynamic inverse. With the feedback controller, the stability of the plant can be guaranteed. On the other hand, for the indirect inverse NN controller, typically a double NNs architecture was introduced~\cite{Selmic2000, ZHOU20071062} with one NN for the nonlinear system identification and thus producing the estimation of the system Jacobian, and another one learning the system dynamics inverse. But the structure is complex and will lead to more parameters thus needs more training data. And it took some trials to update the NNs and drive the tracking error converged. Another problem of the inverse NN controller reported in these works is that the weights of the NNs should be retrained for a new task, so that it is only suitable for repetitive tasks. Gaussian Processes~\cite{Williams2009}
and Locally Weighted Projection Regression (LWPR) were also reported to approximate robot inverse dynamics.

Controller designs combining NNs and ILC for trajectory tracking have also been investigated, with NNs generating control signals for updating the input iteratively~\cite{Chow2000, Zuo2010, Chien2002, Wang2004, PATAN201713402, HAN20183641}.
In these works, the weights of NNs were updated iteratively and the update law needs to be carefully designed to guarantee the bounded tracking error. The resulting controller structure is complex and on-line iterations are necessary to achieve required accuracy so it only applies to repetitive tasks. Moreover, the controllers were tested only in simulation or on relatively simpler physical systems than robots.

Overall,~\cite{Li2017} appears closest to our work. Li et al.~\cite{Li2017} proposed an NN-based control architecture with an NN pre-cascaded to a quadrotor under a feedback controller. The NN produces a reference input to the feedback control system with which the quadrotor can track a given unseen desired trajectory accurately without online iteration. In~\cite{Zhou2018ral} they extend the framework to non-minimum phase systems. Compared to their approach, our method is different in the following aspects: 1) we apply the trained NNs to an articulated industrial robot, which has very different dynamics from the quadrotors; 2) the work in~\cite{Li2017, Zhou2018ral} collected large amount of training data directly from the physical system, which is not practical for rigid industrial robots. Instead, we use simulation-based gradient descent ILC to collect training data for the NNs, which is safer and more time-efficient. In addition, we transfer the knowledge from simulation to the physical system, and also explore the potential of transferring the knowledge from one robot model to another model in simulation; 3) they took a sequence of the future desired states and the current state feedback of the quadrotor as input of the NN, and the NN produced only one reference input for current time step. On the contrary, the input of our NNs is only composed of the desired trajectory. Specifically, the input of our NNs contains both the past desired states and future desired states and the NNs will output the reference input for multiple future time steps.


In this paper, we use multilayer feedforward NNs trained by ILC off-line to directly approximate the inverse dynamics of the robot, without computing the system Jacobian. The reasons for choosing ILC include that it can achieve almost perfect trajectory tracking, and the implementation of ILC is facilitated by a dynamical simulator so that the training data collection process is convenient. A thorough review of ILC is in~\cite{Bristow2006}. Similar to ILC, DDP is another gradient-based optimization algorithm to search for local optimal control inputs. For implementation on nonlinear systems, DDP algorithm iteratively updates the control input by applying the linear quadratic regulator (LQR) to the linearized system about the current trajectory, which makes it impractical in general to solve DDP at runtime. ILC, on the other hand, updates the control input iteratively based on the errors of the previous iterations, which can be obtained either through a dynamic system model or the real plant. With a high-fidelity simulator, the iteration steps can be executed autonomously, as described in Section~\ref{methodology}. However, the significant drawback of these algorithms is that the knowledge learned for one specific task is not
transferable to other tasks. Thus, ILC still has to be re-implemented through multiple iterations
before achieving high tracking precision for a new trajectory. Our approach, on the other hand, combines the advantages of both NNs and ILC, which allows off-line training ahead of time, and the trained model can generalize well to arbitrarily reasonable trajectories and different robot models without any adaption process. In addition, since there is no closed-loop action, the system is always stable.


Compared to state-of-the-art of NN-based controller design, the contributions of this paper are summarized as follows.
\begin{itemize}
    \item We design a gradient descent ILC law for MIMO nonlinear systems.
    \item Our system is stable and we do not need to design the update law of the NNs but just use the common backpropagation. The training of NNs is off-line and the trained NNs effectively approximate the dynamic inversion, and can generalize well to unseen trajectories without on-line iteration.
    \item We apply the noncausual scheme to train the NN to accommodate the inverse of the strictly proper nonlinear system.
    \item We train the NNs by simulation data and then transfer the NNs to the physical robot to compensate for the reality gap.
    \item Most of the state-of-the-art works for NN-based control only test the feasibility of their approaches by simulation or on simple physical systems like mass-spring-damper. Whereas we test the feasibility of our approach by intensive simulation and experimental trials with a 6 DOF nonlinear industrial robot.
    \item We explore the generality of the trained NNs on two different robot models in simulation and it demonstrates that the tracking performance of these robots can improve a lot without any adaption of the NNs. Thus, the trained NNs have the potential to be served as a general purpose network.
\end{itemize}

\section{Problem Statement}
\label{problem statement}

\begin{figure}[tb]
    \centering
    \includegraphics[width=3.4in]{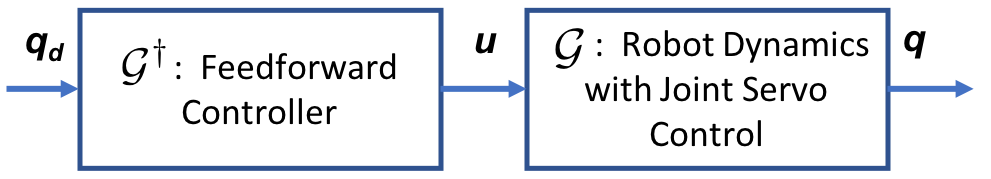}
    \caption{Robot dynamics with closed loop joint servo control.}
    \label{fig:innerloop}
\end{figure}

Consider an $n$-dof robot arm under closed loop joint servo control with input $u\in\rr n$, typically either the joint velocity command $\dot q_c$ or the joint position command $q_c$, and the output $q\in\rr n$, the measured joint position. The inner loop dynamics, $\calG$, is a sampled-data nonlinear dynamical system possibly with communication transport delay and quantization effect. Our goal is to find a feedforward compensator $\calG^\dagger$ (as shown in Fig.~\ref{fig:innerloop}) such that for a given desired joint trajectory $q_d$, the feedforward control $u=\calG^\dagger q_d$ would result in asymptotic tracking: $q=\calG u\to q_d $, as $t\to \infty$. 

\begin{figure}[tb]
    \centering
    \includegraphics[width=0.45\textwidth]{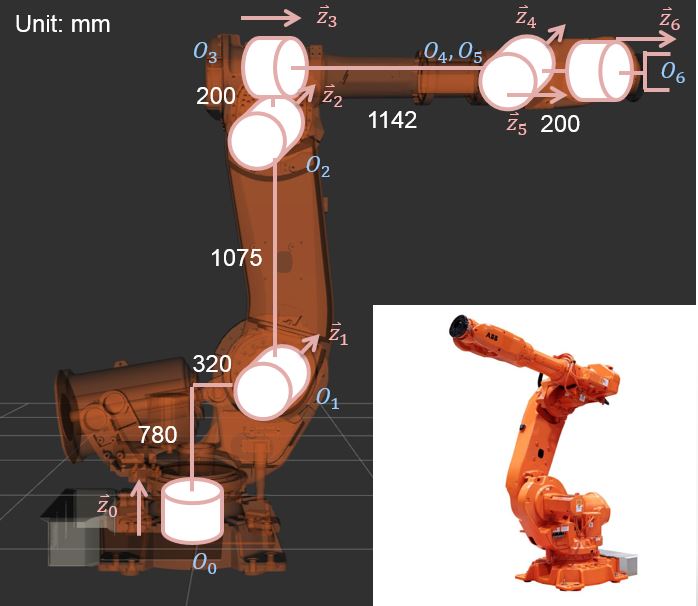}
    \caption{ABB IRB 6640-180 robot (inset) and the kinematic parameters.}
    \label{fig:kinematics}
\end{figure}

We will use ABB~IRB6640-180 robot, shown in Fig.~\ref{fig:kinematics}, as a test platform. IRB6640-180 is a robot manipulator with six revolute joints, 180~kg load capacity and maximum reach of 2.55~m. It has high static repeatability (0.07~mm) and path repeatability (1.06~mm)~\cite{irb6640_product_manual}. In this paper, we assume the robot internal dynamics and delays are deterministic and repeatable, which is reasonable considering the high accuracy and repeatability of the robot. The ABB controller has an optional {\em Externally Guided Motion} (EGM) module which provides an external interface of commanded joint angles, $u=q_c$, and joint angle measurements at 4~ms rate. 
ABB also offers a high fidelity dynamical simulator, RobotStudio~\cite{robotstudio}, with the same EGM feature included.
We use this simulator to implement ILC and in turn generate a large amount of training data for ${\calG}^\dagger$ implemented as a multi-layer NN.

\begin{figure}[tb]
\centering
\setlength{\unitlength}{0.012500in}%
\includegraphics[width=0.45\textwidth]{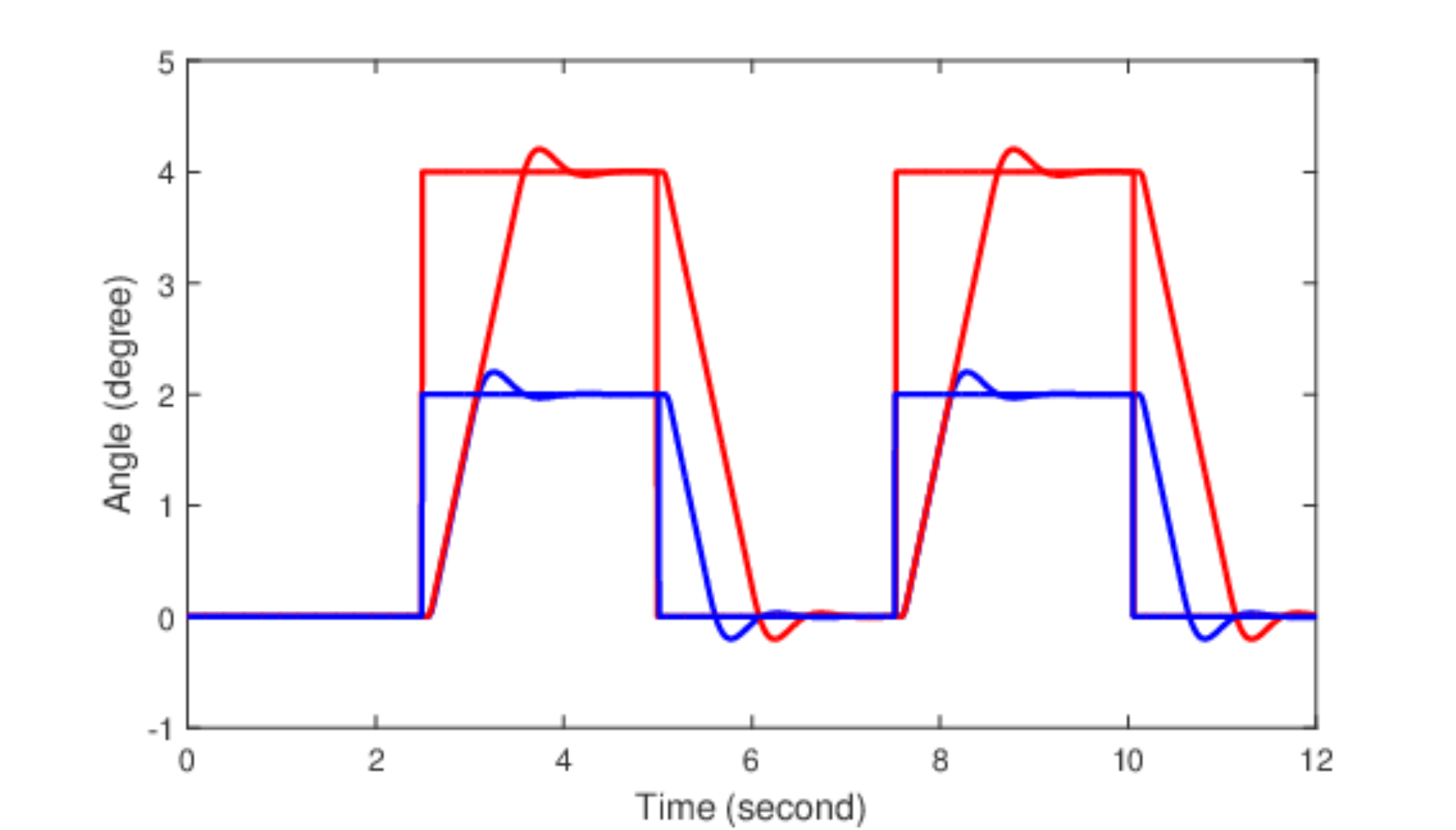}
\caption{Multiple step responses with different input amplitudes (2 and 4 degrees) and at two different configurations denoting the nonlinearity of $\calG$.}
\label{fig:stepresp} 
\end{figure}

EGM sends the commanded joint angle $q_c\in{\mathbb R}^6$ to the low level servo controller as a setpoint, and reads the actual joint angle vector, $q\in{\mathbb R}^6$, measured by the encoders. Ideally, $\calG$ should be the identity matrix. But because of the robot dynamics and the inner control loop, $\calG$ is a nonlinear dynamical system.   Fig.~\ref{fig:stepresp} shows step responses of EGM under multiple input amplitudes (2 and 4 degrees) and at different configurations. The responses show classic under-damped behavior, but the nonlinearity (reduced gain and slower dynamics at higher input amplitude) and time delays are clearly visible.

\section{Methodology}
\label{methodology}

Though $\calG u$ may be implemented in simulation and physical experiments, finding $\calG^\dagger$ is challenging because $\calG$ is difficult to characterize analytically.  Our approach to the approximation of $\calG^\dagger$ does not require the explicit knowledge of $\calG$ and only uses $\calG u$ for a specific input $u$. The key steps include:
\begin{enumerate}
    \item {\em Model-free Gradient Descent-based Iterative Refinement}: For a specific desired trajectory $q_d$, we approximate $\calG$ by a linear time invariant system $G(s)$.  The gradient descent direction $\calG^* (q-q_d)$ ($\calG^*$ is the adjoint of $\calG$), may be approximated by $G^*(s)(q-q_d)=G^T(-s)(q-q_d)$. As shown in \cite{Potsaid2007}, this can be obtained using $\calG u$ for a suitably chosen $u$. 
    \item {\em Approximation of Dynamical System using a Feedforward Neural Network}:
    The first step generates multiple $(q_d,u)$ trajectory pairs corresponding to $u=\calG^\dagger q_d$.  We would like to approximate $\calG^\dagger$ by a feedfoward neural network and train the weights using $(q_d,u)$ from iterative refinement. To enable a feedforward net to approximate a dynamical system, we use a batch of $q_d$ to generate a batch of $u$.  Furthermore, the $q_d$ batch is shifted with respect to $u$ to remove the effect of transient response.
    \item {\em Improvement of Neural Network based on Physical Experiments}: We perform the first two steps using a dynamical simulator of $\calG$ (in our case, RobotStudio). When the trained NN approximation of $\calG^\dagger$ is applied to the physical system, tracking error may increase due to the discrepancy between the simulator and the physical robot.  Instead of repeating the process all over again using the actual robot, we will just retrain the output layer weights of the NN. 
\end{enumerate}    
The rest of this section will describe these steps in greater details.





\subsection{Iterative Refinement}

\begin{figure}[tb]
\centering
\setlength{\unitlength}{0.012500in}%
\includegraphics[width=0.48\textwidth]{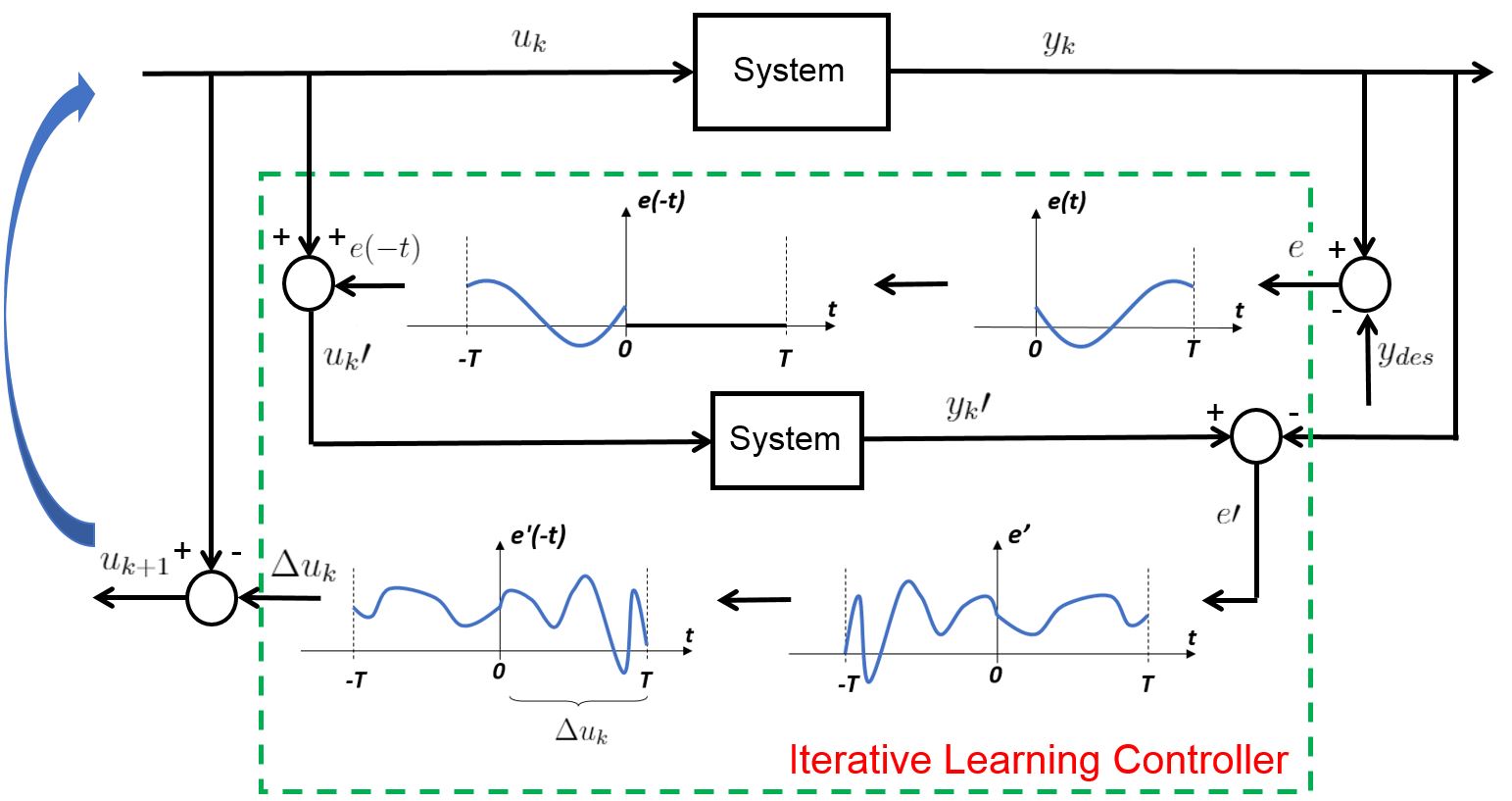}
\caption{Block diagram of the gradient descent-based ILC algorithm.}
\label{fig:ilc} 
\end{figure}

Consider a single-input/single-output (SISO) linear time invariant system with transfer function $G(s)$, input $u$ and output $y$. We use underline of a signal to denote the trajectory of the signal over $[0,T]$ for a given $T$. 
Given a desired trajectory $\ydbar$, the goal is to find an input trajectory $\ubar$ so that the $\ell_2$-norm of the output error, $\norm{\ybar - \ydbar}$, is minimized. 
Given an initial guess of the input trajectory $\ubar\sup0$, we may use gradient descent to update $\ubar$:
\begin{equation}
\ubar\sup{k+1} = \ubar\sup{k}-\alpha_k G^*(s) \eybar
\label{eq:ILC}
\end{equation}
where $G^*(s)$ is the adjoint of $G(s)$, 
$\eybar:=(\ybar\sup k-\ydbar)$, and $\ybar\sup k = G(s) \ubar \sup k$. 
%
The step size $\alpha_k$ may be selected using line search to ensure the maximum rate of descent. 
Let the state space parameters of a strictly proper $G(s)$ be $(A,B,C)$, i.e., $G(s)=C(sI-A)\inv B$.  Then $G^*(s)= G\tr(-s) = B\tr (-sI-A\tr)\inv C\tr$.  Since $A$ is stable, $G\tr(-s)$ would be unstable and we cannot directly compute $G^*(s) \eybar$.  For a stable computation of the gradient descent direction, we use the technique described in \cite{Potsaid2007}.  We propagate backward in time with time-reversed $\eybar$ to stably compute the time-reversed gradient direction.  The result is then reversed in time to obtain the gradient descent direction forward in time. 
The key insight here is to perform the gradient generation using $\calG\eybar$ instead of an analytical model of $\calG$. As a result, no explicit model information is needed in the iterative input update.
The process of one iteration of gradient descent based on $\calG\eybar$ is shown in Fig.~\ref{fig:ilc} and described below:
\begin{itemize}
\item[(a)] Apply $\ubar\sup k$ to the system at a specific configuration and obtain the output $\ybar\sup k = \calG \ubar\sup k$ and corresponding tracking error $\eybar\sup k = \ybar\sup k - \ydbar\sup k$.
\item[(b)] Time-reverse $e_y(t)$ to $e_{y_R}(t)=e_y(-t)$. 
\item[(c)] 
Define the augmented input ${u'}\sup k(t) = u\sup k(t) +e_{y_R}(t)$.
Let ${y'}\sup k=\calG {u'}\sup k(t)$ with the system at the same initial configuration as in step (a).
\item[(d)] Compute $e'(t)= {y'}\sup k(t) - y\sup k(t)$, and reverse it in time again  to obtain the gradient direction $(\calG\eybar)(t) = e'(-t)$.
\end{itemize}

The above iterative refinement algorithm is easily applied to a single robot joint with $\calG $ given by RobotStudio.  The first guess of $\ubar\sup 0$ is simply chosen to be the desired output trajectory $\ydbar$. 
We may also fit the response of RobotStudio using a third order system (an underdamped second order system cascaded with a low pass filter) within the linear regime of the response:
\begin{equation}
G(s) = \frac{a}{s+a}\times\frac{\omega^2}{s^2+2\zeta\omega s + \omega^2}.
\label{eq:transfer_function}
\end{equation}
Within the linear regime, $G(s)$ may be used instead of $\calG$ to speed up the iteration (as RobotStudio is more time consuming than computing a linear response).

\begin{figure}[tb]
   \centering
   \begin{subfigure}[b]{0.49\textwidth}
        \includegraphics[width=\textwidth]{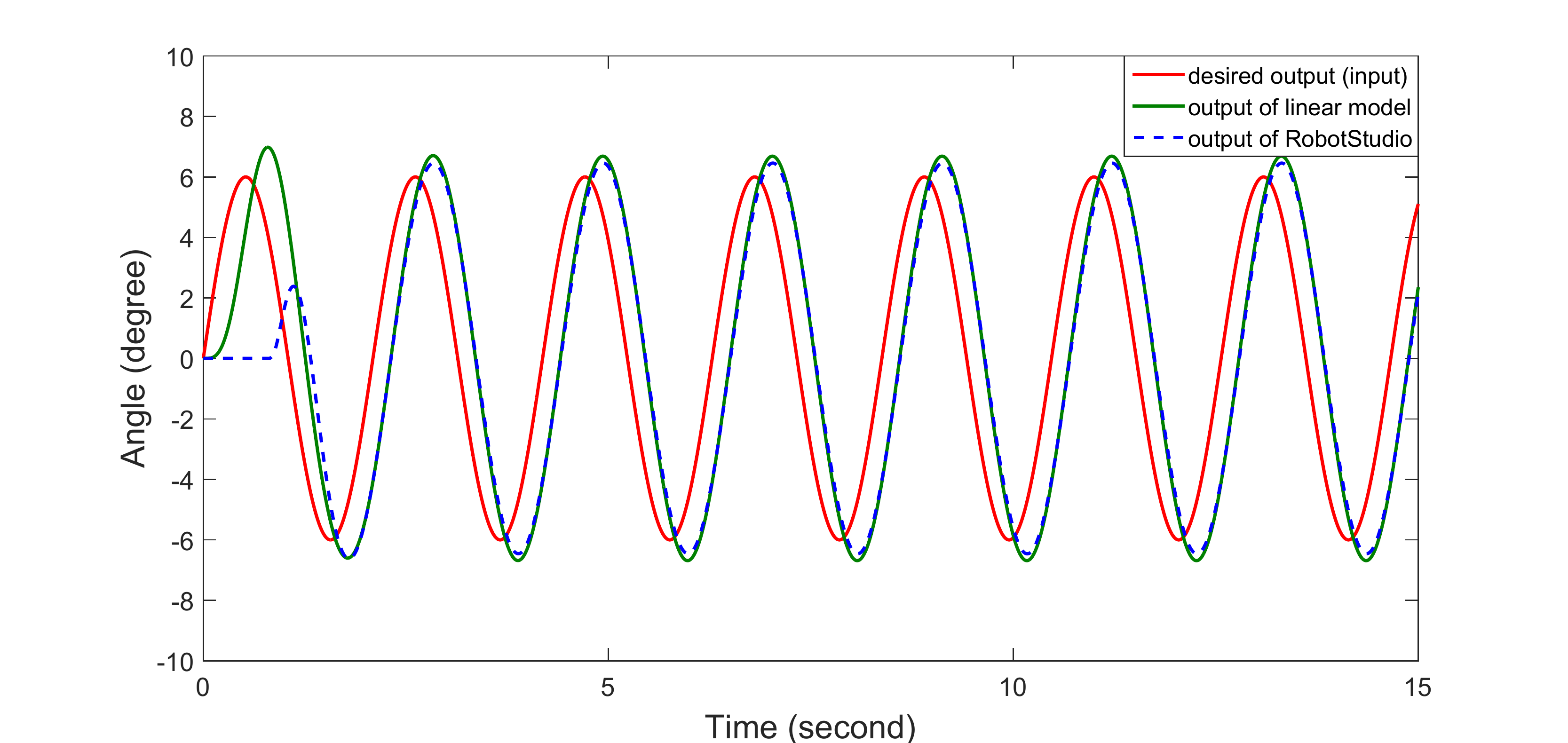}
        \caption{Desired output (also
the input into the inner loop), RobotStudio output, and linear model output.}
        \label{fig:tracking_no_learing}
    \end{subfigure}
    \hfill
    \begin{subfigure}[b]{0.5\textwidth}
        \includegraphics[width=\textwidth]{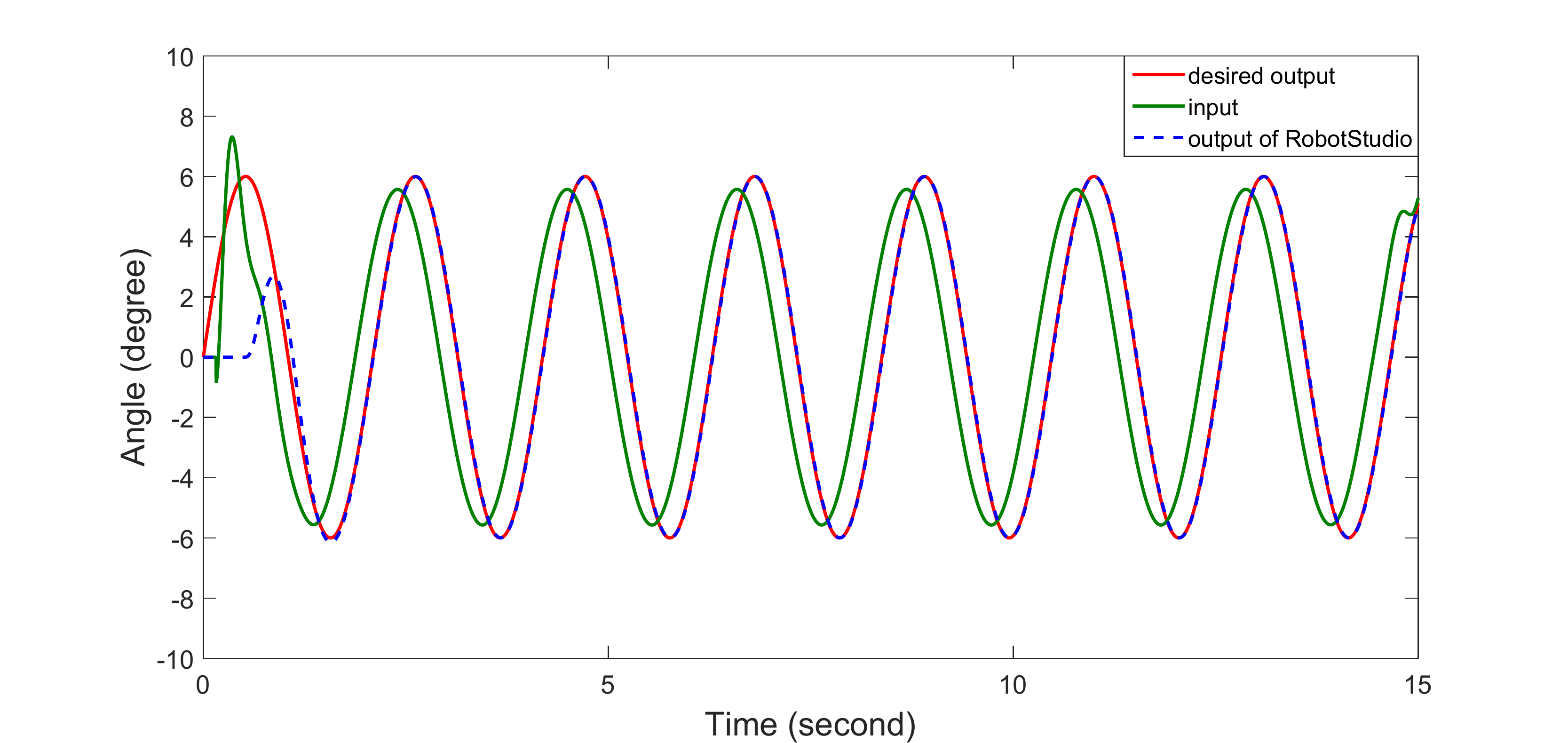}
        \caption{Desired output, RobotStudio output, and modified input based on iterative refinement.}
        \label{fig:tracking_with_learing}
    \end{subfigure}
   \caption{Trajectory tracking improvement with iterative refinement after 8 iterations. The desired trajectory is a sinusoid with $\omega$ = 3 rad/sec. }
   \label{fig:sin_tracking}
   \vspace{-5pt}
\end{figure}

Table~\ref{single_joint_path} compares the tracking error $\norm{\ybar-\ydbar}$ in RobotStudio, with using RobotStudio or the linear model to generate the gradient descent direction.  As the amplitude and frequency of the desired output is still in the linear regime, both approaches work well. 
Fig.~\ref{fig:sin_tracking} shows the trajectory tracking improvement of a sinusoidal path.  Fig.~\ref{fig:tracking_no_learing} shows the uncompensated case (i.e., $\ybar\sup 0 =\calG \ydbar$).  RobotStudio output shows an initial deadtime and transient, but matches well with the linear response afterwards. Because of the RobotStudio dynamics, there is a significant tracking error in both amplitude and phase. Fig.~\ref{fig:tracking_with_learing} shows the input $\ubar$ and output $\ybar$ after 8 iterations. After the initial transient, $\ybar$ and $\ydbar$ are almost indistinguishable.  The input $\ubar$ shows amplitude reduction and phase lead compared to $\ydbar$ (also $\ubar\sup 0$) to compensate for the amplitude gain and phase lag of RobotStudio for the input $\ydbar$.


\begin{table}[tb]
\caption{Comparison of the $\ell_2$ norm of tracking errors for a single joint with gradient calculated based on RobotStudio and the fitted linear model. The initial tracking error is 31.6230$^{\circ}$.}
\label{single_joint_path}
\begin{center}
\begin{tabular}{|c|c|c|}
\hline
 Iterations \# & Tracking Error & Tracking Error \\
 & Using RobotStudio ($^{\circ}$) & Using Linear Model ($^{\circ}$) \\
\hline
1 & 26.0699 & 25.0585 \\
2 & 21.0735 & 20.0049 \\
3 & 17.1372 & 15.9724 \\
4 & 13.6738 & 12.7540 \\
5 & 11.2821 & 10.1849 \\
\hline
\end{tabular}
\end{center}
\end{table}


The inner loop control for industrial robots typically tightly regulates each joint, and the system is almost diagonal as seen from the outer loop. 
Therefore, the single-axis algorithm may be directly extended to the multi-axis case by using the same iterative refinement procedure applied to each axis separately. 

We have developed a Python interface for automated iterative refinement  with optimal learning rate $\alpha_k$ at each iteration. This code is available at: \htmladdnormallink{https://github.com/rpiRobotics/RobotStudio\_EGM\_ILC}{https://github.com/rpiRobotics/RobotStudio\_EGM\_ILC}. 



\subsection{Feedforward Neural Network}

Despite the significantly improved tracking performance, iterative refinement finds $u=\calG^\dagger q_d$ only for a specific $q_d$.
It is therefore not suitable for run-time implementation. 
Our aim is to obtain an approximation of $\calG^\dagger$ directly by learning from a set of $(u,q_d)$ obtained from iterative refinement.
To this end, we use multi-layer NNs to represent $\calG^\dagger$. 
Because the decoupled nature of $\calG$ (command of joint $i$ only affects the motion of joint $i$), we have $n$ separate NNs, one for each joint. 
We train these NNs from scratch using a large number of sample trajectories covering the robot workspace, and the corresponding required inputs are obtained by iterative refinement.

We hypothesize that $\calG$ may be represented as a nonlinear ARMA model.  We use a feedforward net to approximate this system similar to \cite{noriega1998direct}. At a given time $t$, the input of the NN is a segment of the desired joint trajectory $\qdbar(t):=\{q_d(\tau): \tau\in[t-T,t+T]\}$.  The output is also chosen as a segment of $u$: $\ubar:=\{u(\tau): \tau\in[t,t+T]\}$. Updating a segment of the outer loop control instead of a single time sample reduces computation by avoiding frequent invocation of the NN. The choice of the future segment of $\qdbar$ is to allow approximation of noncausal behavior needed in inverting nonminimum phase or strictly proper $\calG$. 

We determine the NN architectures, particularly the number of hidden layers and the number of nodes in each layer, by experiments. The selection of $T$ (identically, the input dimension of NNs) should guarantee that the input of NNs contains enough information for the NNs to model the dynamics, and at the same time should not be too large or it will require a lot more training data. By testing, we choose $T$ to correspond to 25 samples (in the case of EGM with 4~ms sampling period, $T=0.1$~s). The input and output layers therefore have 50 and 25 nodes, respectively. Table~\ref{nn architecture selection} lists part of example results of final mean squared testing error on unseen testing data by using different combinations of the number of hidden layers and the number of nodes of each hidden layer.

\begin{table}[tb]
\caption{Final mean squared testing error of different NN architectures.}
\label{nn architecture selection}
\begin{center}
\begin{tabular}{|c|c|c|}
\hline
 Number of & Number of Nodes & Final Testing Error ($^{\circ}$) \\ Hidden Layers
 &  of Each Layer & (Mean of Three Trials) \\
\hline
2 & 50/100 100/100 & 2.623 2.145 \\
2 & 100/150 100/200 & 2.474 2.545 \\
3 & 100/100/100 100/150/100 & 2.390 2.486 \\
4 & 100/100/100/50 &  2.450\\
\hline
\end{tabular}
\end{center}
\end{table}

After testing, we use a fully-connected network including two hidden layers,
and each hidden layer contains 100 nodes. To traverse the workspace of the robot as much as possible, for tracking each trajectory the robot starts at a random configuration and each trajectory is composed of 3000 samples, or 12~s). Each  trajectory can therefore provide many training segments.


We use a large amount of desired trajectories, including sinusoids, sigmoids and trapezoids to train the NNs in \texttt{TensorFlow} by using backpropagation~\cite{Rumelhart1986} to minimize the loss between the output of NNs and the expected output over a randomly selected batch of training data (with batch size of 256) in each training iteration. Using trajectories with different velocity and acceleration profiles, we sample both linear and nonlinear regions of EGM. As commonly used in practice, we use \texttt{AdamOptimizer} to train the NNs, $L_2$ norm regularization to avoid overfitting, and rectified linear unit (ReLU) activation to introduce nonlinearity for the NNs. 


\subsection{Transfer Learning}

\begin{figure}[tb]
   \centering
   \begin{subfigure}[b]{0.49\textwidth}
        \includegraphics[width=\textwidth]{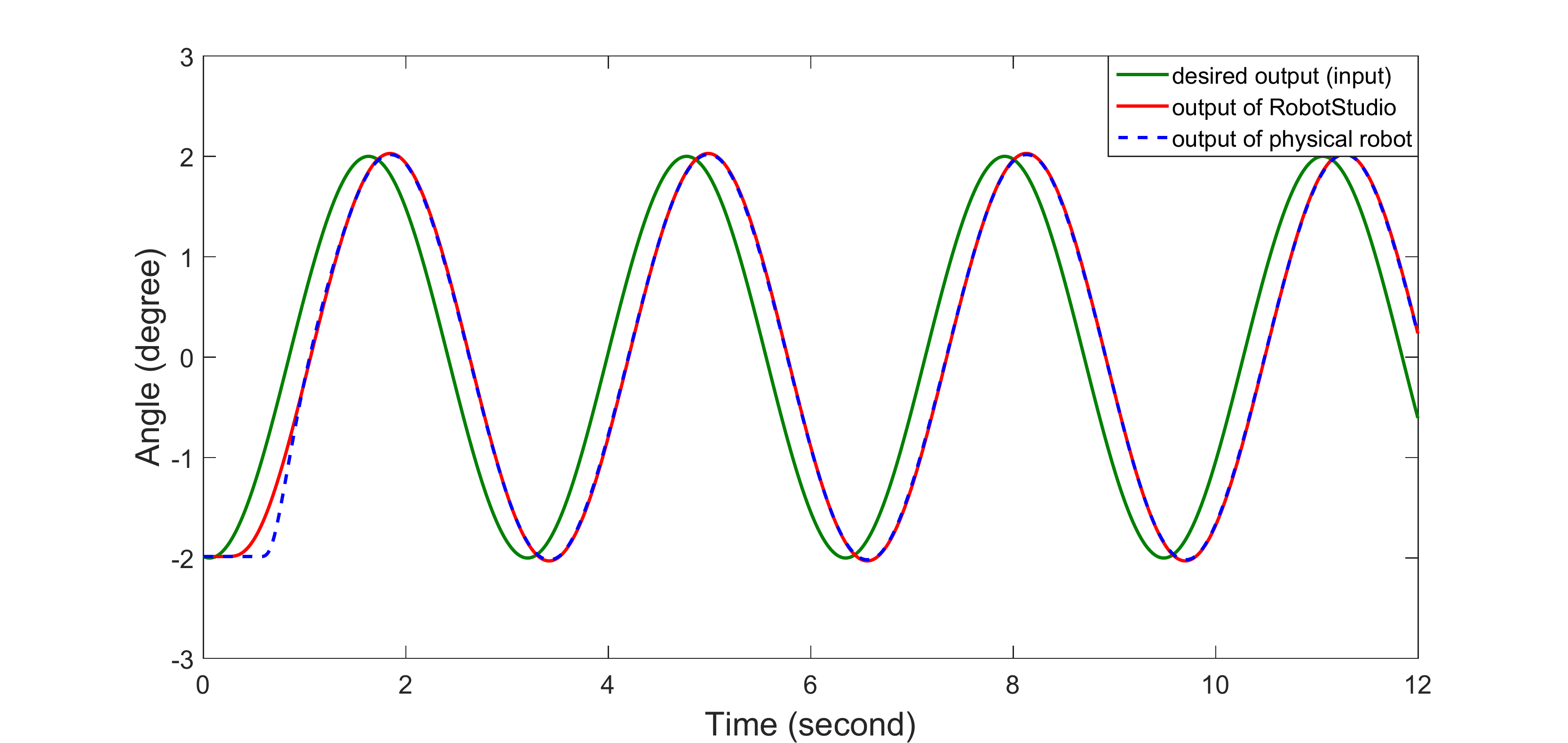}
        \caption{Desired output, physical robot output, and RobotStudio output.}
        \label{fig:sin_mag2_omega2}
    \end{subfigure}
    \hfill
    \begin{subfigure}[b]{0.5\textwidth}
        \includegraphics[width=\textwidth]{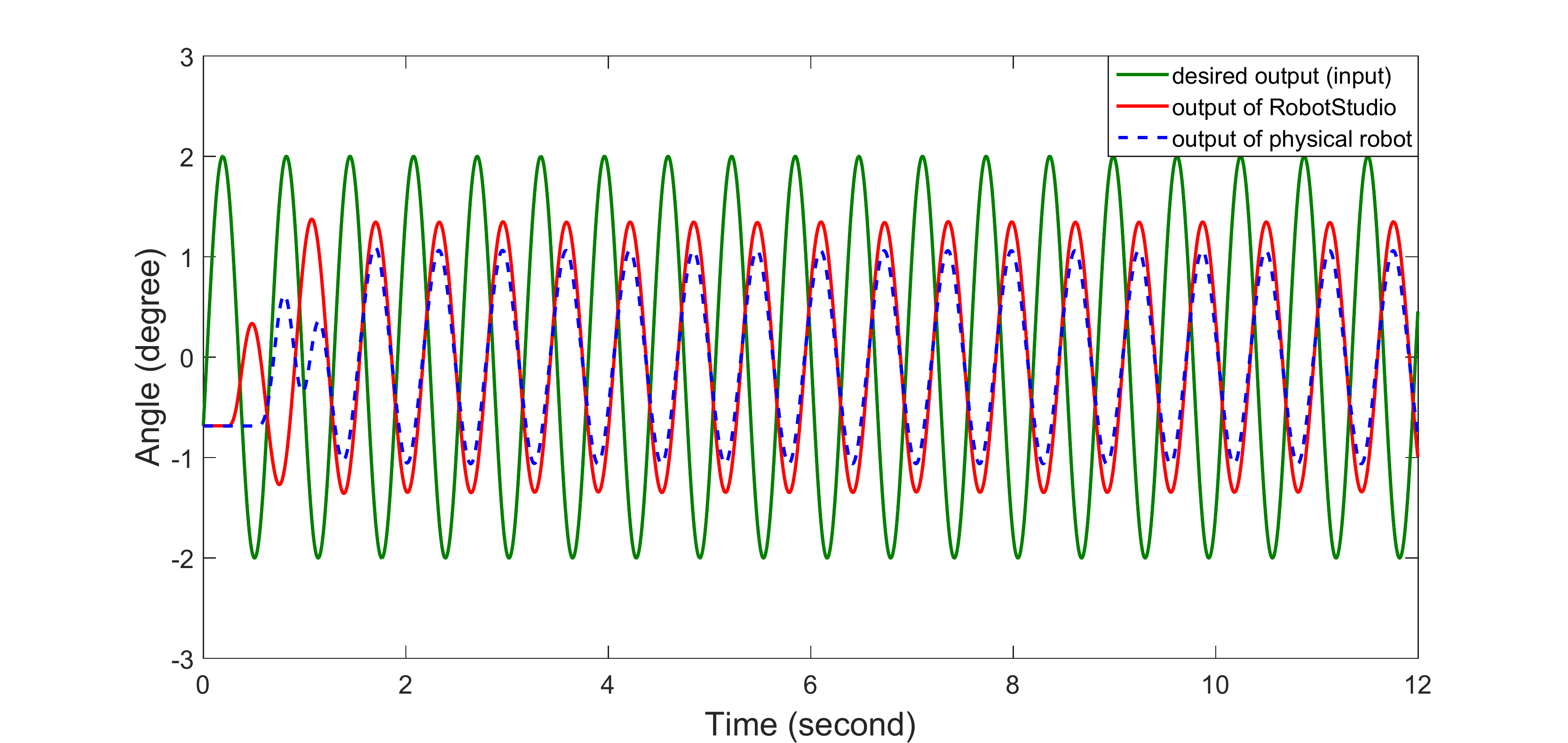}
        \caption{Desired output, physical robot output, and RobotStudio output.}
        \label{fig:sin_mag2_omega10}
    \end{subfigure}
   \caption{Comparison of responses between RobotStudio (red curve) and the physical robot (blue curve).  Desired inputs are sinusoids with amplitude 2$^\circ$ and angular frequency $\omega$ 2~rad/sec and 10~rad/sec.}
   \label{fig:comparison}
   \vspace{-5pt}
\end{figure}

Reality gap exists between RobotStudio and the physical robot due to model inaccuracies and load (mounted grippers and cameras) on the physical robot. RobotStudio can capture the dynamics of the physical robot very well for a trajectory with low velocity, as in Fig.~\ref{fig:sin_mag2_omega2}. However, for trajectories with large velocity and/or acceleration, discrepancies between RobotStudio and physical robot responses become apparent  (mainly in the output magnitude) as shown in Fig.~\ref{fig:sin_mag2_omega10}. 

We use additional iterative learning results on the physical robot to train only the output weights of the NNs to narrow the reality gap between the simulator and the real plant. We apply the iterative refinement procedure on the physical robot for 20 different trajectories and use the results  to fine-tune the NNs.  The trained NNs are essentially regarded as a feature extractor. We fix the parameters of the first two layers of the trained NNs, and only retrain the small amount of parameters of the output layer with collected training data.  As only the amplitudes of the response show discrepancy, it is reasonable to fine tune only the output layer using limited amount of data from the physical robot.


\section{Results}
\label{results}

\subsection{RobotStudio Simulation Results on IRB6640-180}

\subsubsection{Trapezoidal Motion Profile}

\begin{figure}[tb]
  \begin{subfigure}[b]{0.51\linewidth}
    \centering
    \includegraphics[width=\linewidth]{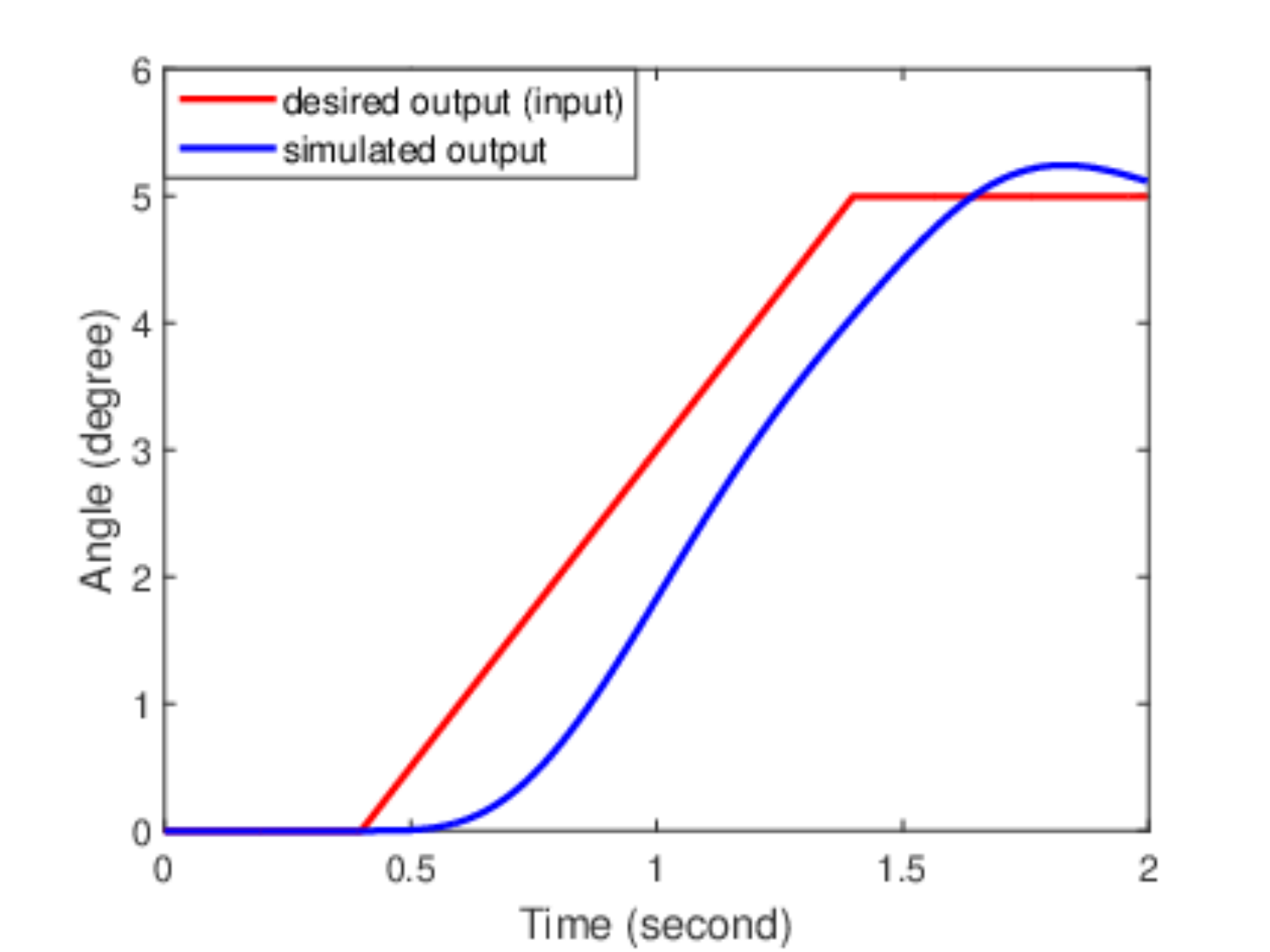}
    \caption{Uncompensated case}
    \label{fig:trapezoid_original}
  \end{subfigure}
  \hspace{-0.5cm}
  \begin{subfigure}[b]{0.51\linewidth}
    \centering
    \includegraphics[width=\linewidth]{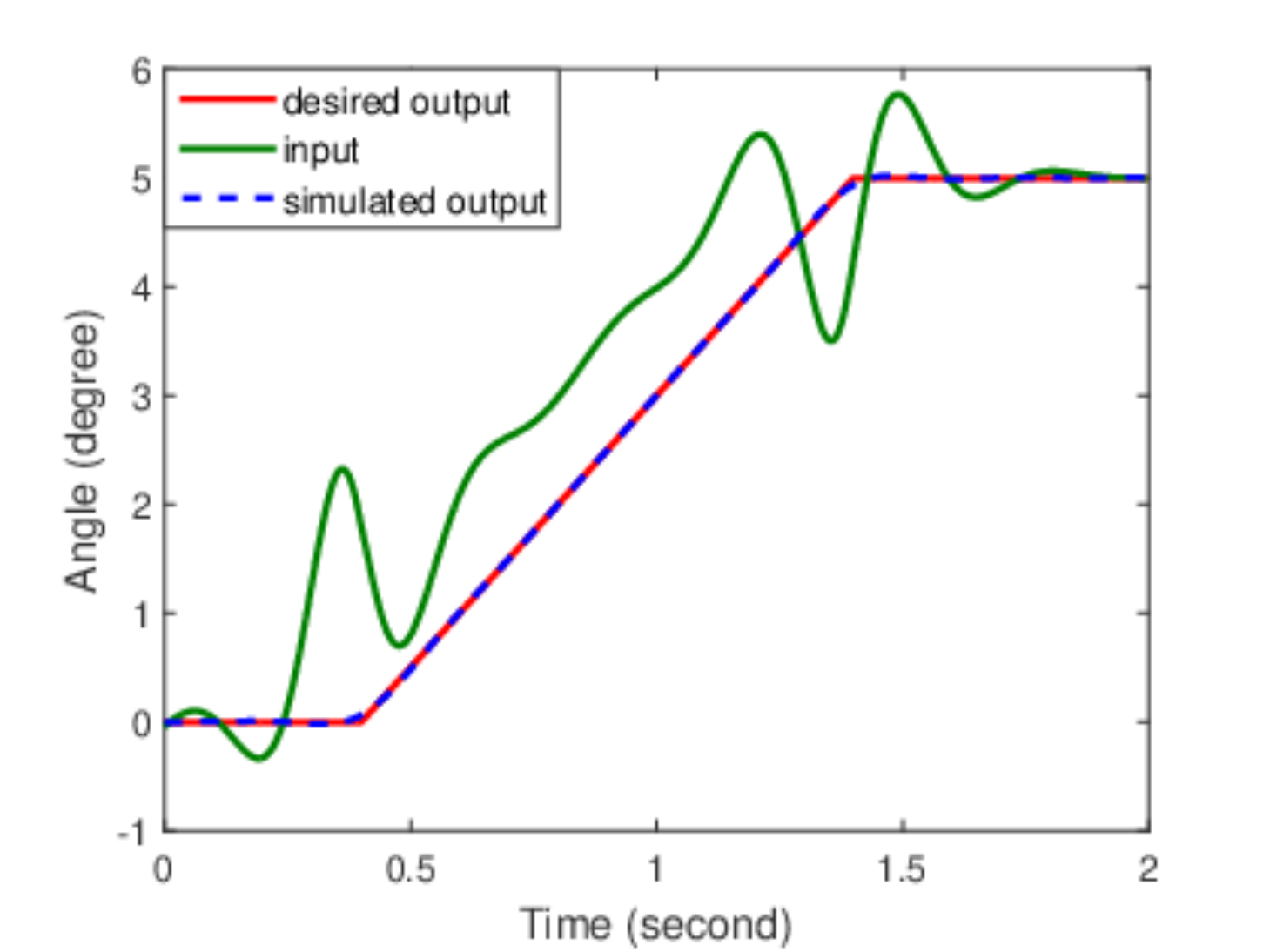}
    \caption{ILC compensated case}
    \label{fig:trapezoid_ilc}
  \end{subfigure}
  \caption{Comparison of tracking performance without and with ILC of a trapezoidal path with high acceleration.}
  \label{fig:trapezoid_iterative}
\end{figure}

Fig.~\ref{fig:trapezoid_original} shows the RobotStudio output of tracking a desired trapezoidal velocity profile with sharp corners (large acceleration at the corner). 
Fig.~\ref{fig:trapezoid_ilc} demonstrates the optimal input obtained by ILC for the trapezoidal path, which shows large input oscillation to achieve the desired tracking. To avoid such undesirable behavior, the desired trajectory should be smooth and have bounded velocity and acceleration such that the optimal input obtained by ILC is reasonable.

\subsubsection{Single Joint Motion Tracking}
\label{Single Joint Motion Tracking}

\begin{figure}[tb]
   \centering
   \begin{subfigure}[b]{0.49\textwidth}
        \includegraphics[width=\textwidth]{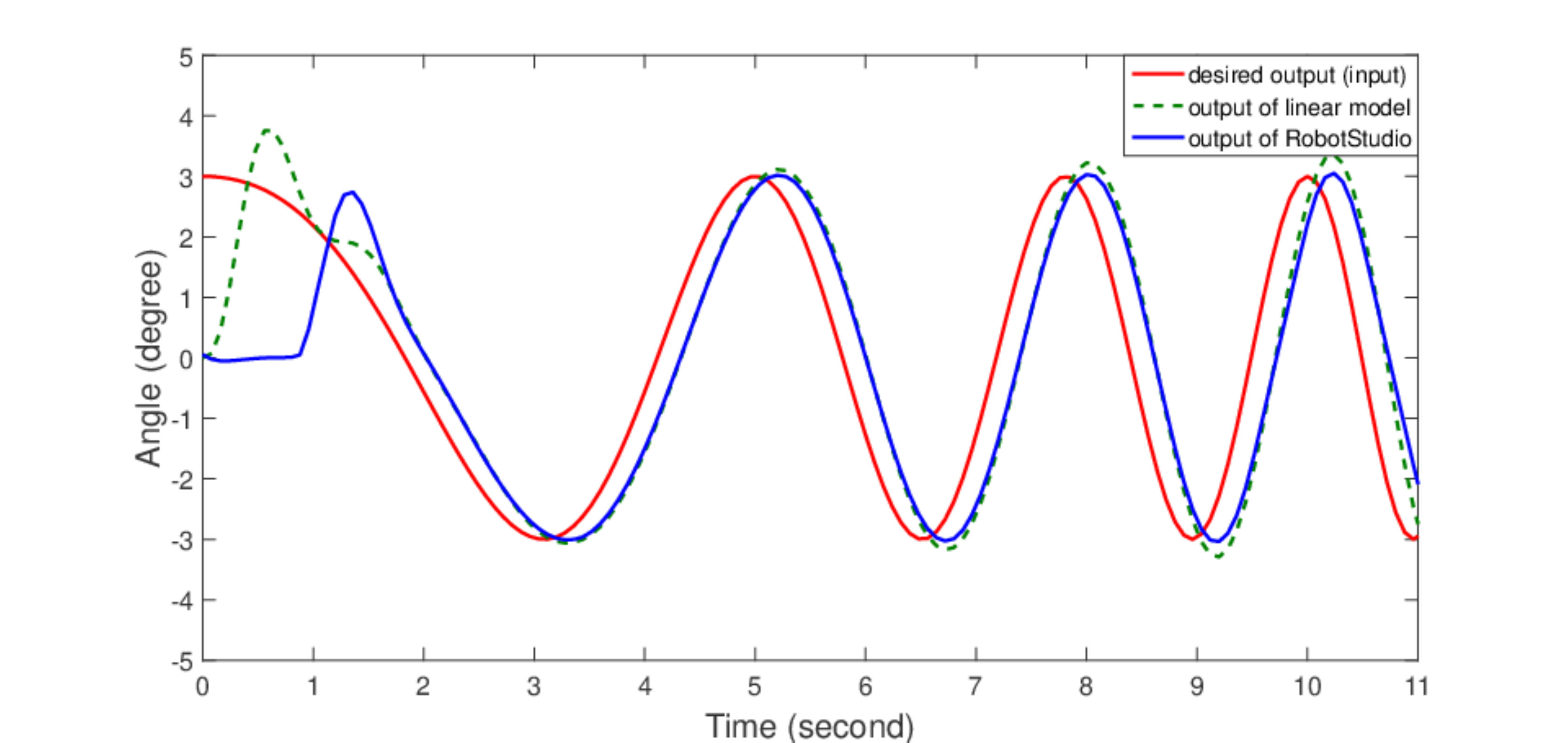}
        \caption{Uncompensated case: desired output (also the input into the inner loop), RobotStudio output, and linear model output.}
        \label{fig:chirp_before}
    \end{subfigure}
    \hfill
    \begin{subfigure}[b]{0.5\textwidth}
        \includegraphics[width=\textwidth]{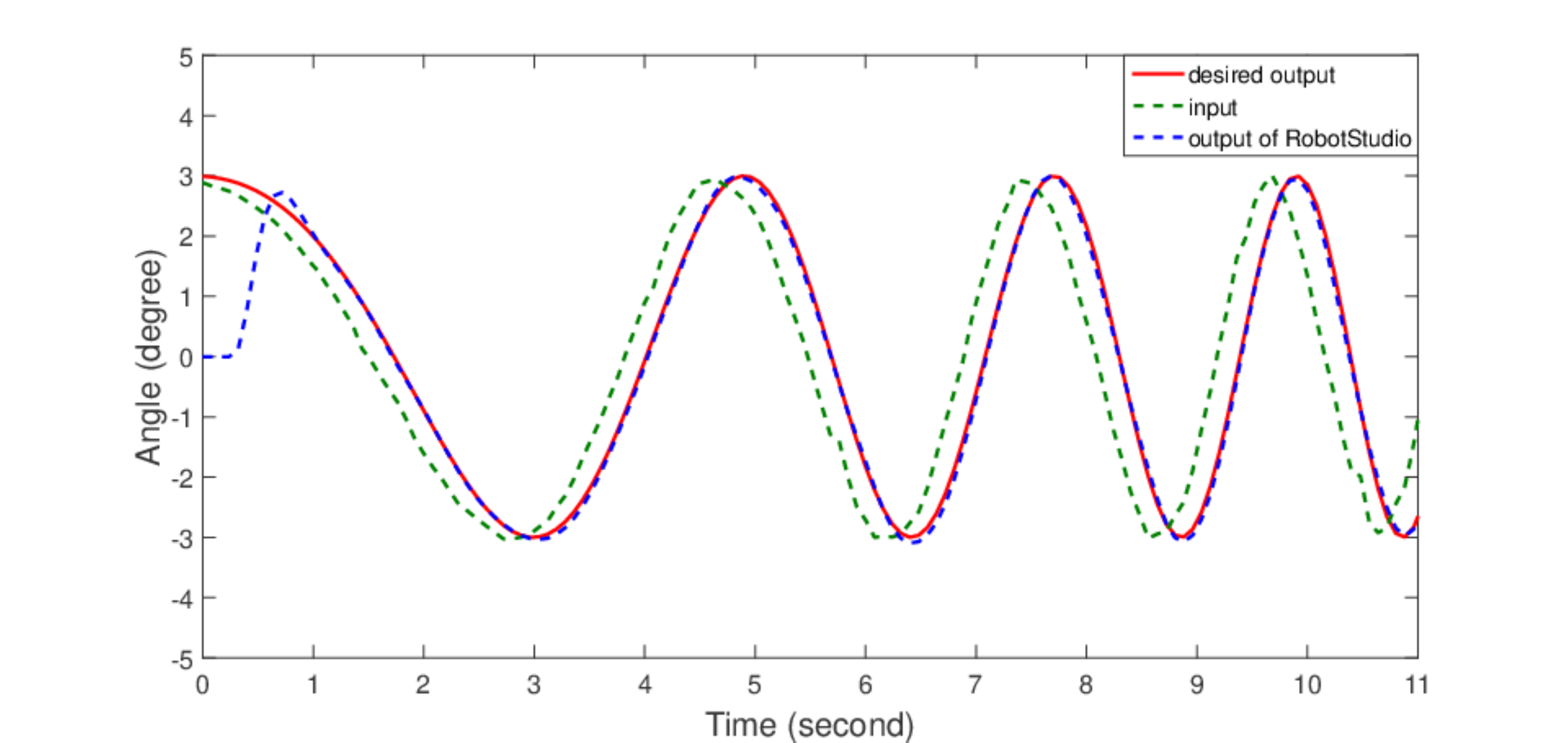}
        \caption{Compensation with  NN:  desired output,  RobotStudio output, and the input generated by the NN.}
        \label{fig:chirp_final}
    \end{subfigure}
   \caption{Comparison of tracking performance without and with the NN compensation for a chirp signal in RobotStudio.}
   \label{fig:chirp}
   \vspace{-5pt}
\end{figure}

As multiple sinusoids have been used in the training of the NN (with discrete angular frequencies ranging from 0.5 to 15 rad/s), we test its generalization ability on a chirp (multi-sinusoids) which is not part of the training set (with continuous angular frequencies ranging from 0.628 to 1.884 rad/s). 
Fig.~\ref{fig:chirp} 
shows the result for joint 1 of the simulated robot with the trained NN. Fig.~\ref{fig:chirp_before} shows 
the uncompensated output with the robot output lagging behind the desired output. The nonlinear effects are denoted by the deviation of the linear model output from the RobotStudio output.
Fig.~\ref{fig:chirp_final} shows that, by compensating for the lag effect and amplitude discrepancies, the command input generated by the NN drives the output to the desired output closely after a brief initial transient.

\subsubsection{Multi-Joint Motion Tracking}

\begin{figure*}[!ht]
   \centering
   \begin{subfigure}[b]{0.32\textwidth}
        \includegraphics[width=\textwidth]{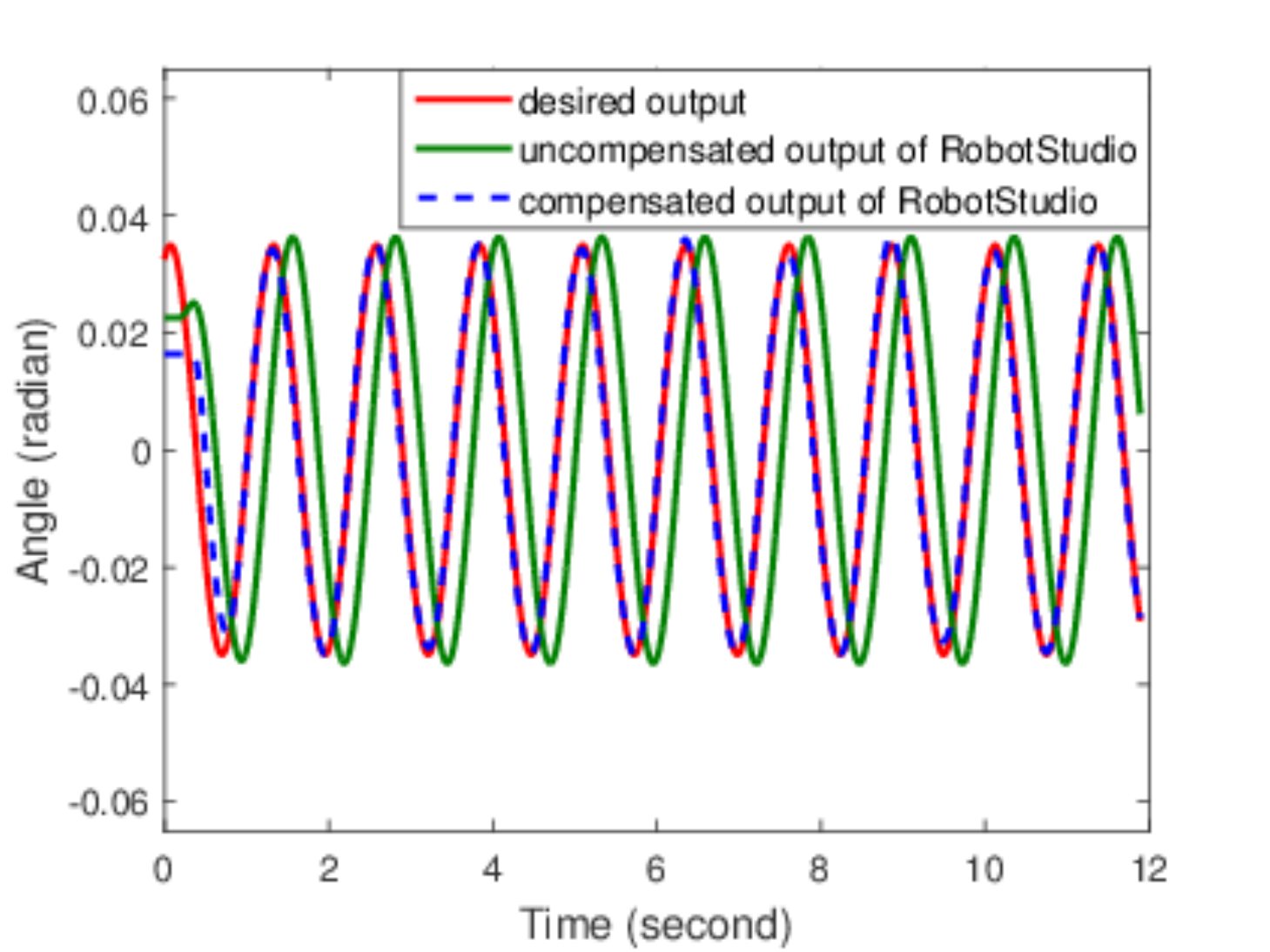}
        \caption{Joint 1}
        \label{fig:final_J1}
    \end{subfigure}
    \begin{subfigure}[b]{0.32\textwidth}
        \includegraphics[width=\textwidth]{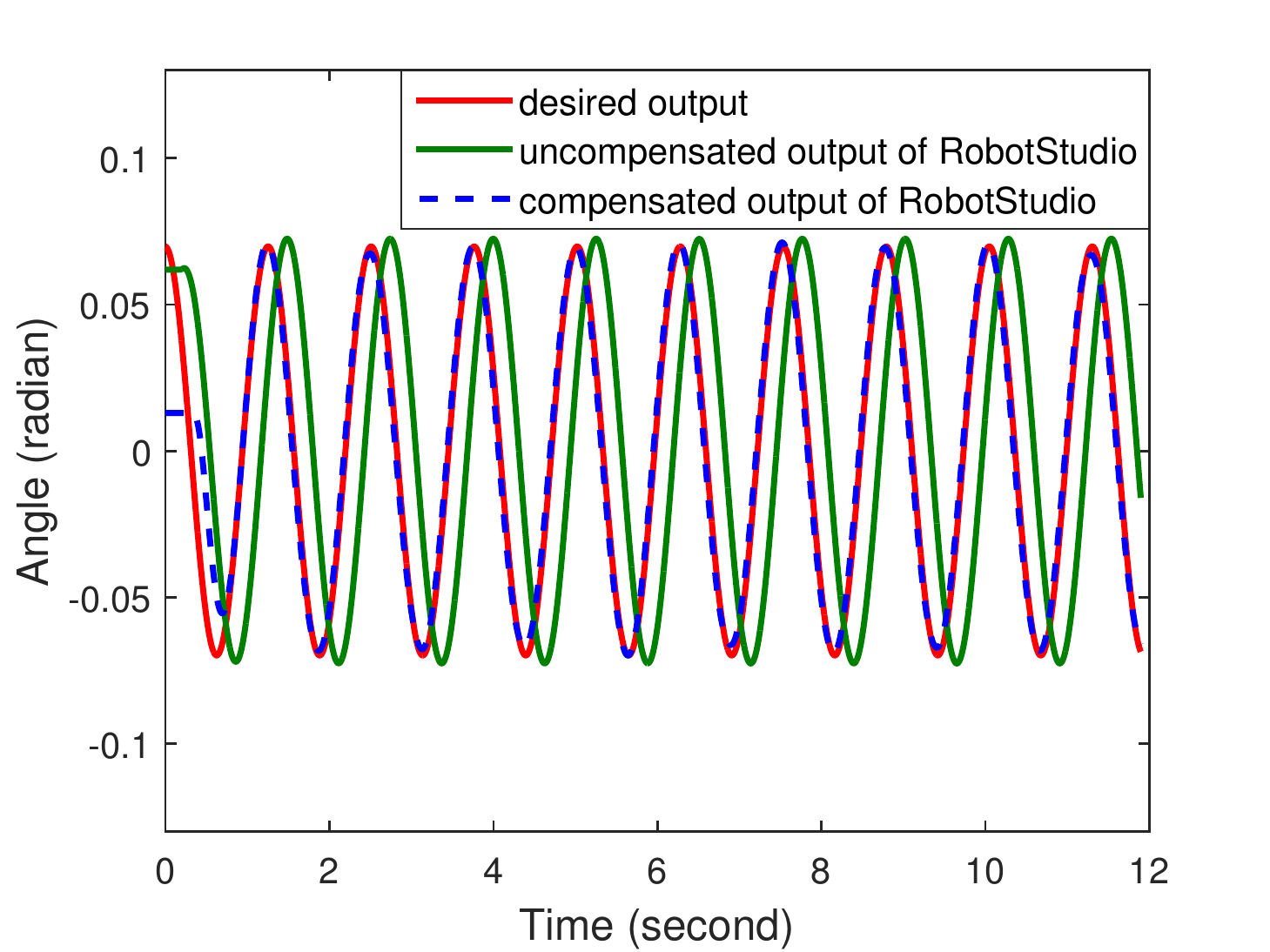}
        \caption{Joint 2}
        \label{fig:final_J2}
    \end{subfigure}    
    \begin{subfigure}[b]{0.32\textwidth}
        \includegraphics[width=\textwidth]{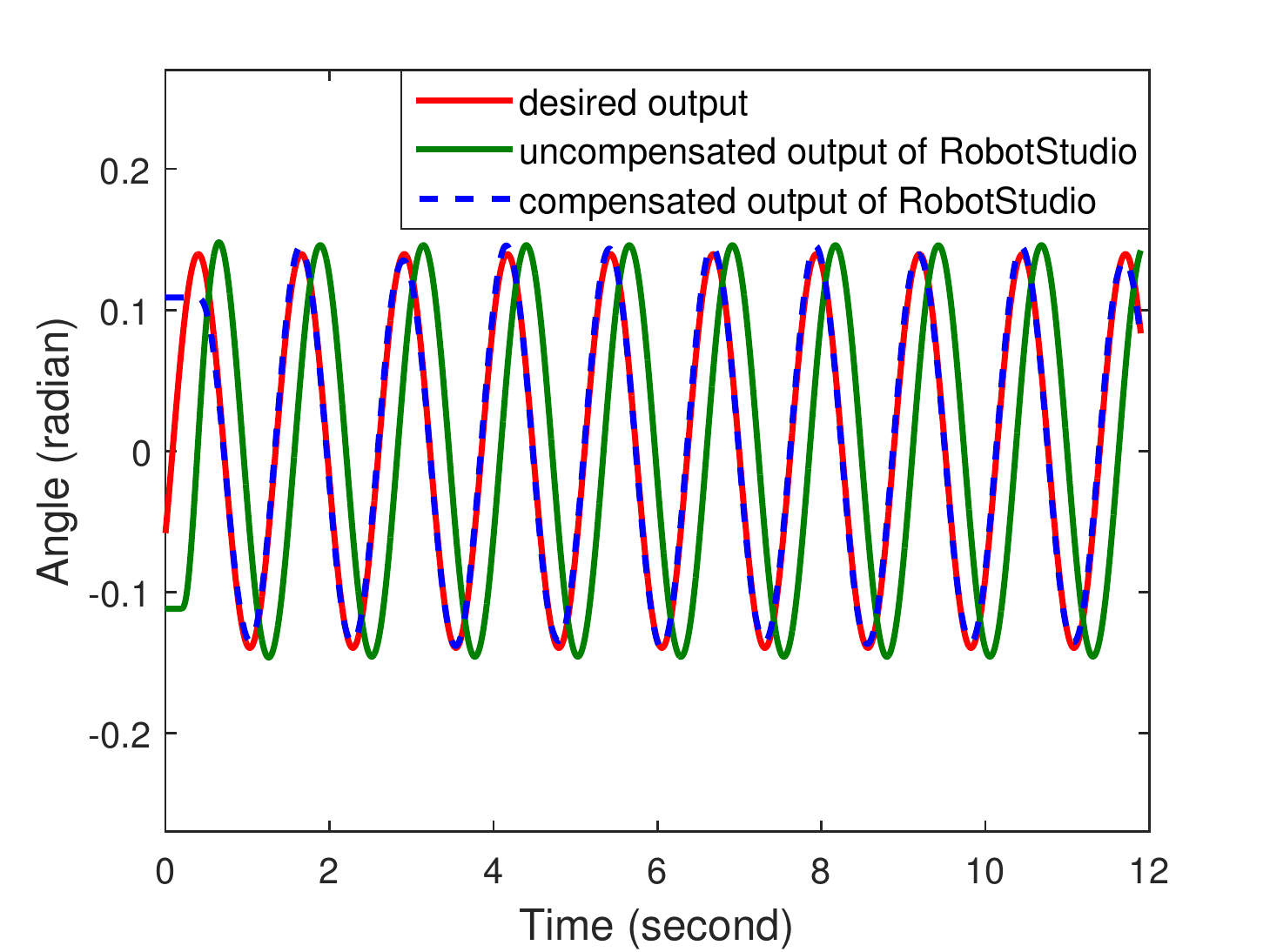}
        \caption{Joint3 }
        \label{fig:final_J3}
    \end{subfigure}    
    \begin{subfigure}[b]{0.32\textwidth}
        \includegraphics[width=\textwidth]{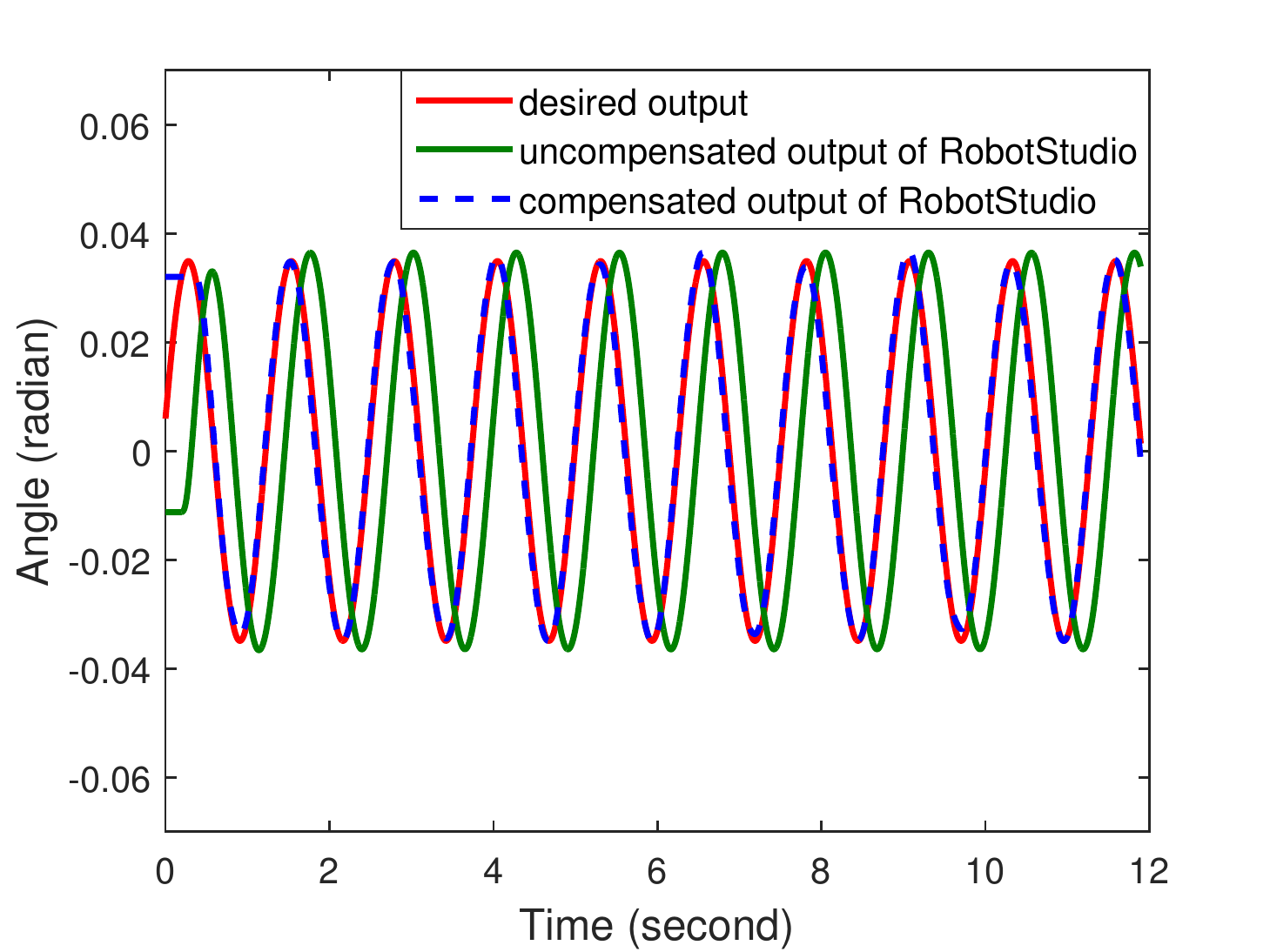}
        \caption{Joint 4}
        \label{fig:final_J4}
    \end{subfigure}    
    \begin{subfigure}[b]{0.32\textwidth}
        \includegraphics[width=\textwidth]{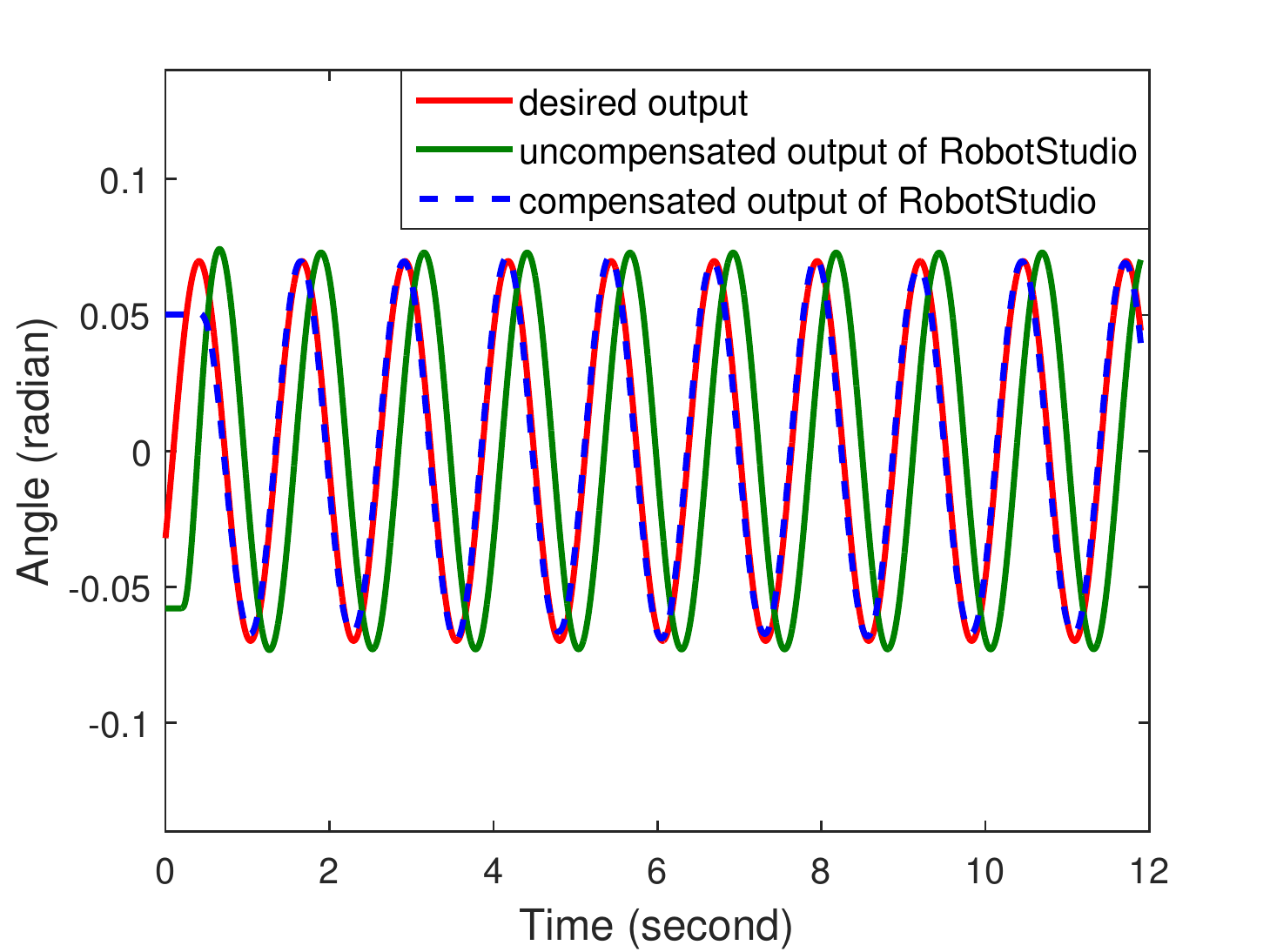}
        \caption{Joint 5}
        \label{fig:final_J5}
    \end{subfigure}    
    \begin{subfigure}[b]{0.32\textwidth}
        \includegraphics[width=\textwidth]{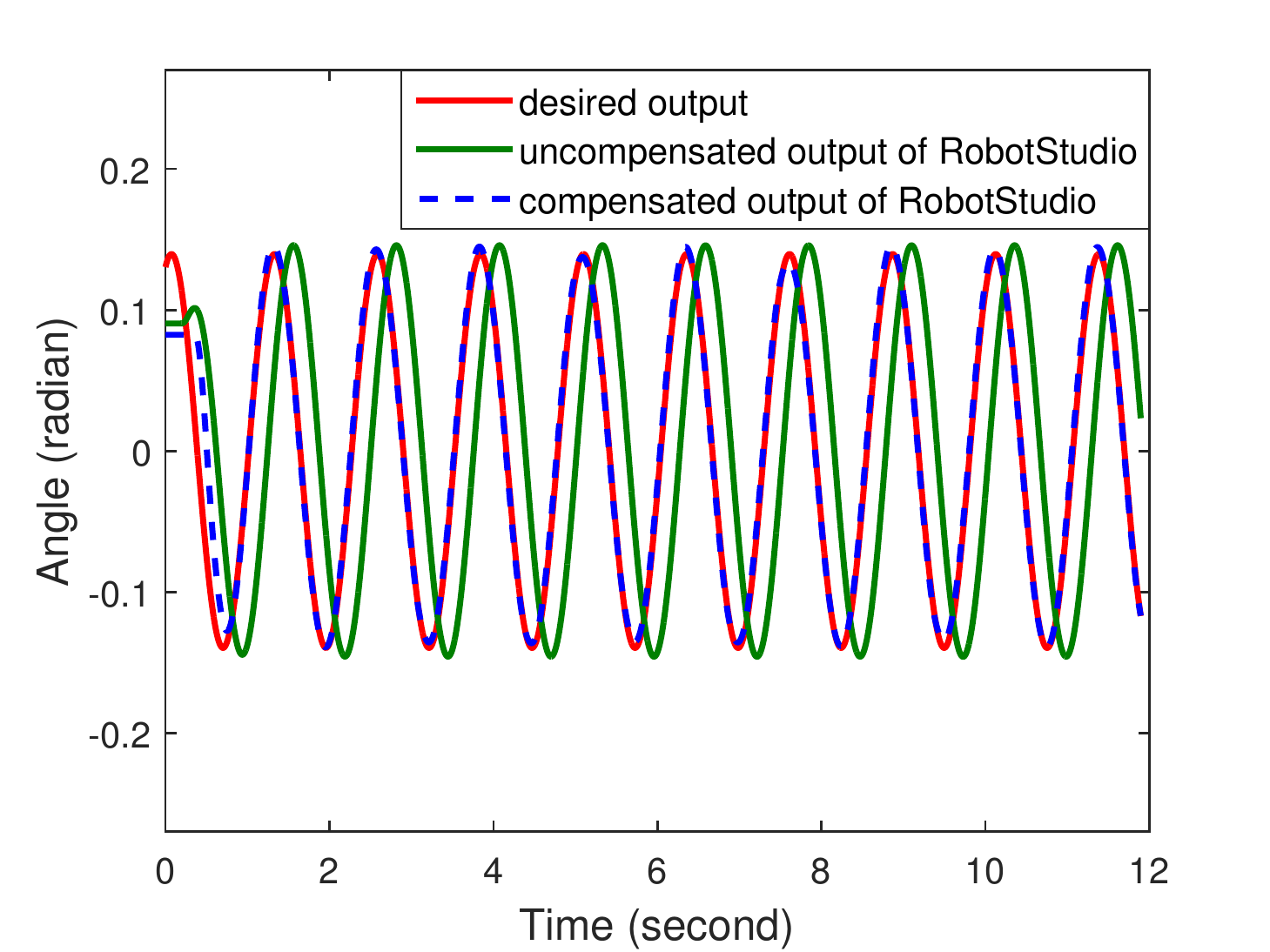}
        \caption{Joint 6}
        \label{fig:final_J6}
    \end{subfigure}
   \caption{Comparison of tracking performance with and without the NNs compensation of sinusoidal joint trajectories in RobotStudio.}
   \label{fig:sin_comparison}
   \vspace{-5pt}
\end{figure*}

Fig.~\ref{fig:sin_comparison} shows the comparison of tracking 6-dimensional random sinusoidal joint trajectories ($\omega$ = 5 rad/s, and with different amplitudes and phases from the training data) without and with NNs compensation. Table~\ref{error_6_joint_path} lists the corresponding tracking errors in terms of $\ell_2$ and $\ell_\infty$ norms (using error vectors at each sampling instant). The $\ell_\infty$ error is calculated by ignoring the initial transient.

\begin{figure*}[tb]
   \centering
   \begin{subfigure}[b]{0.32\textwidth}
        \includegraphics[width=\textwidth]{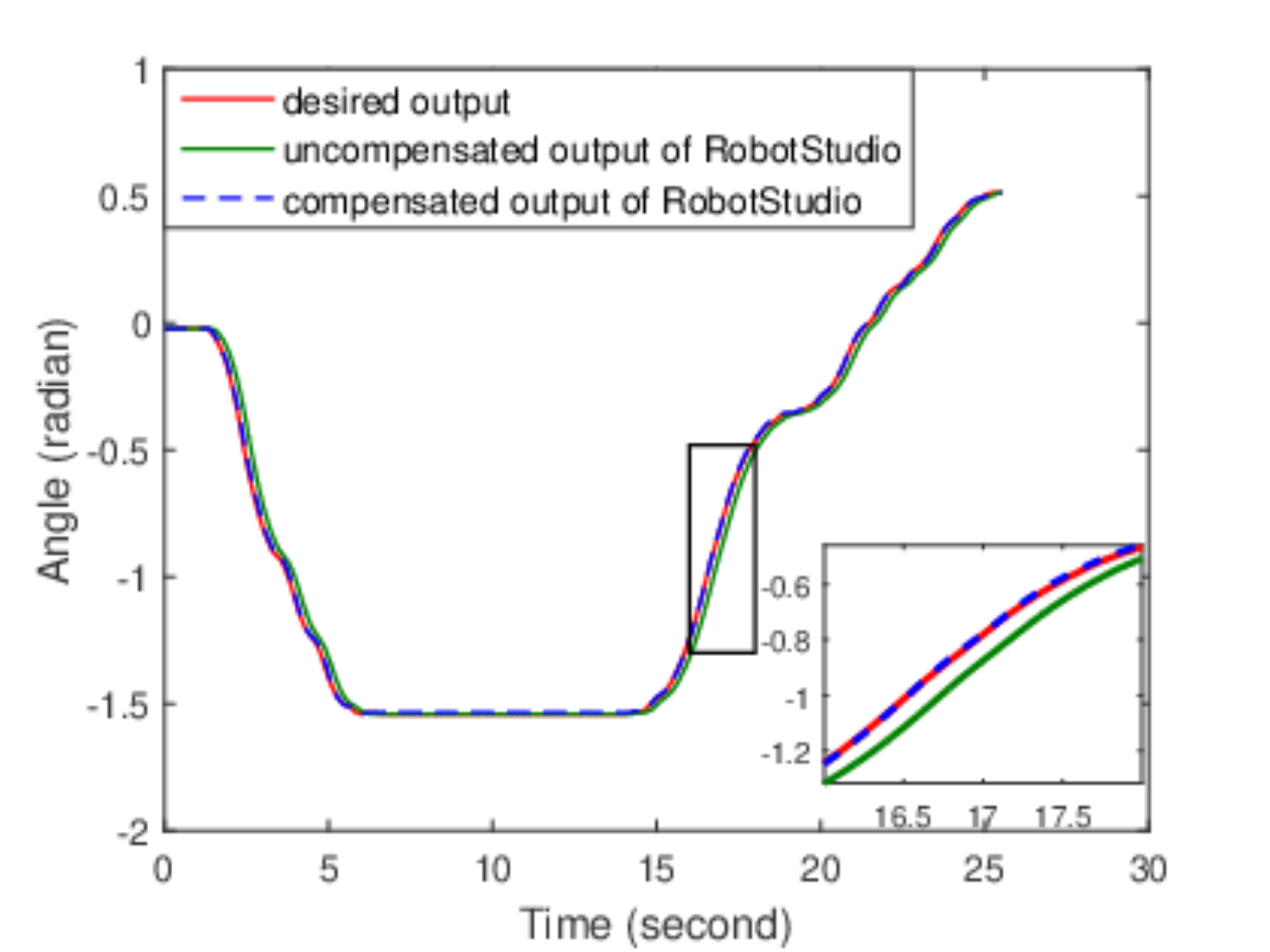}
        \caption{Joint 1}
        \label{fig:J1_RS_moveit_original_final}
    \end{subfigure}
    \begin{subfigure}[b]{0.32\textwidth}
        \includegraphics[width=\textwidth]{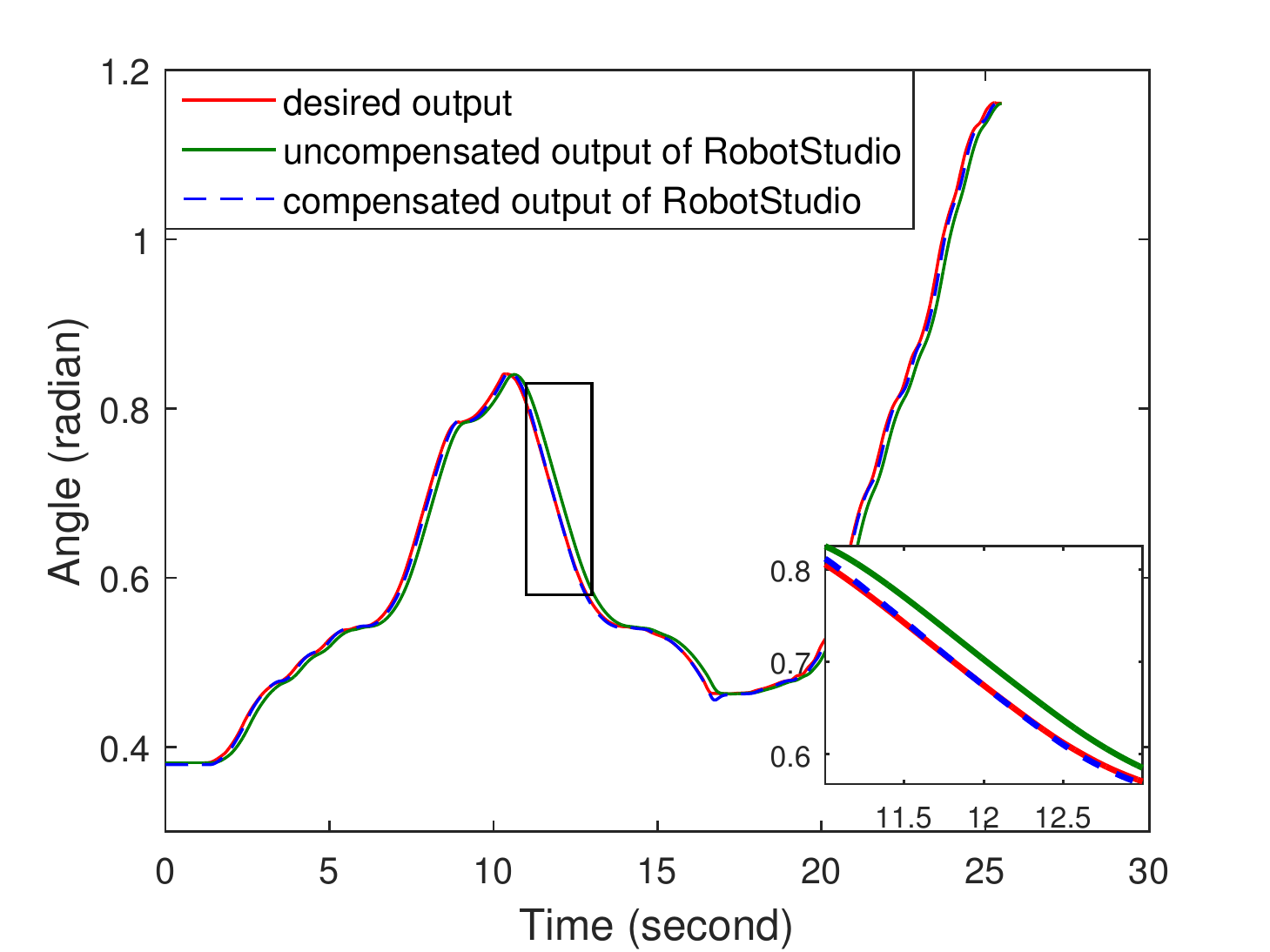}
        \caption{Joint 2}
        \label{fig:J2_RS_moveit_original_final}
    \end{subfigure}    
    \begin{subfigure}[b]{0.32\textwidth}
        \includegraphics[width=\textwidth]{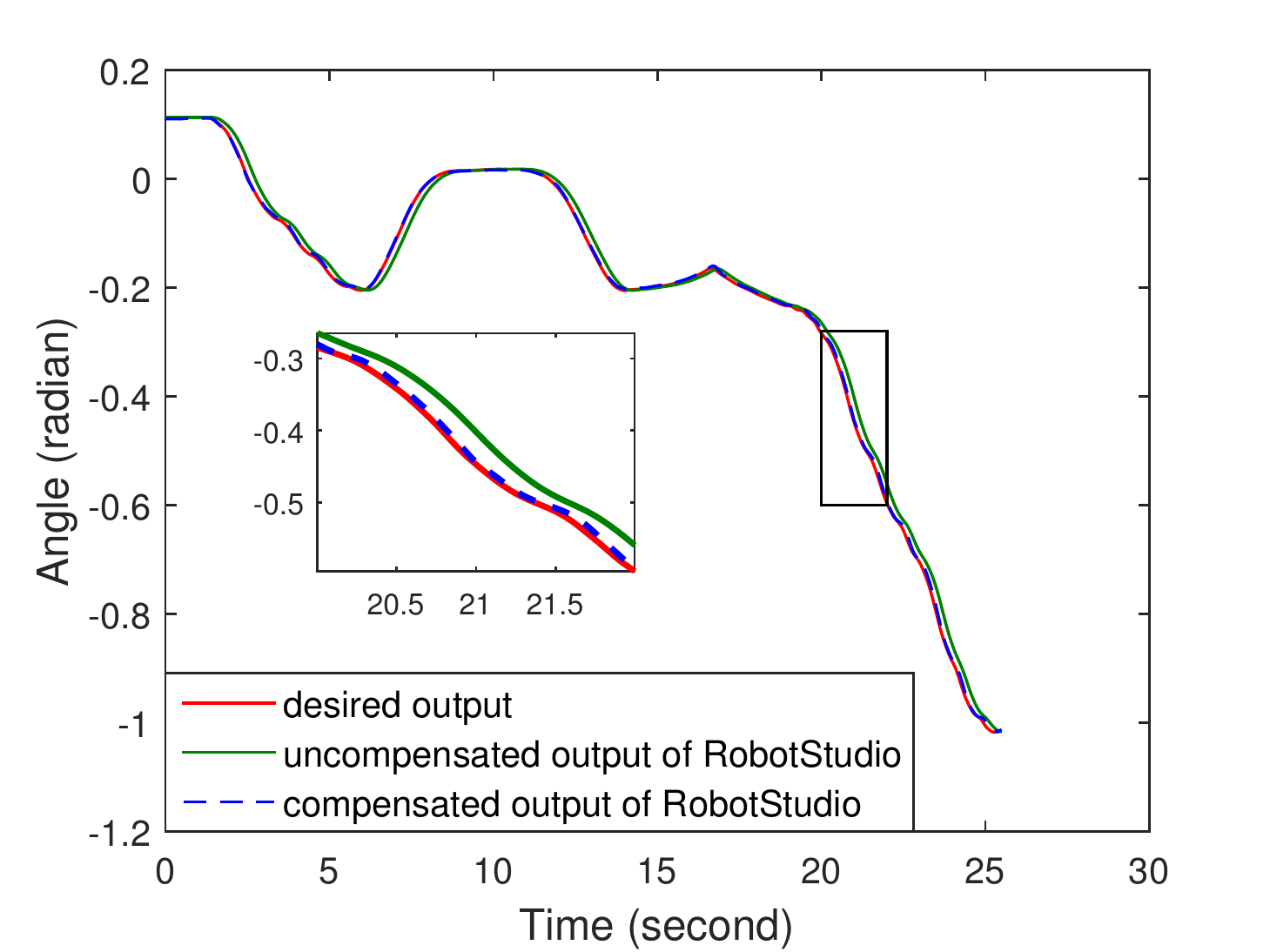}
        \caption{Joint 3}
        \label{fig:J3_RS_moveit_original_final}
    \end{subfigure}    
    \begin{subfigure}[b]{0.32\textwidth}
        \includegraphics[width=\textwidth]{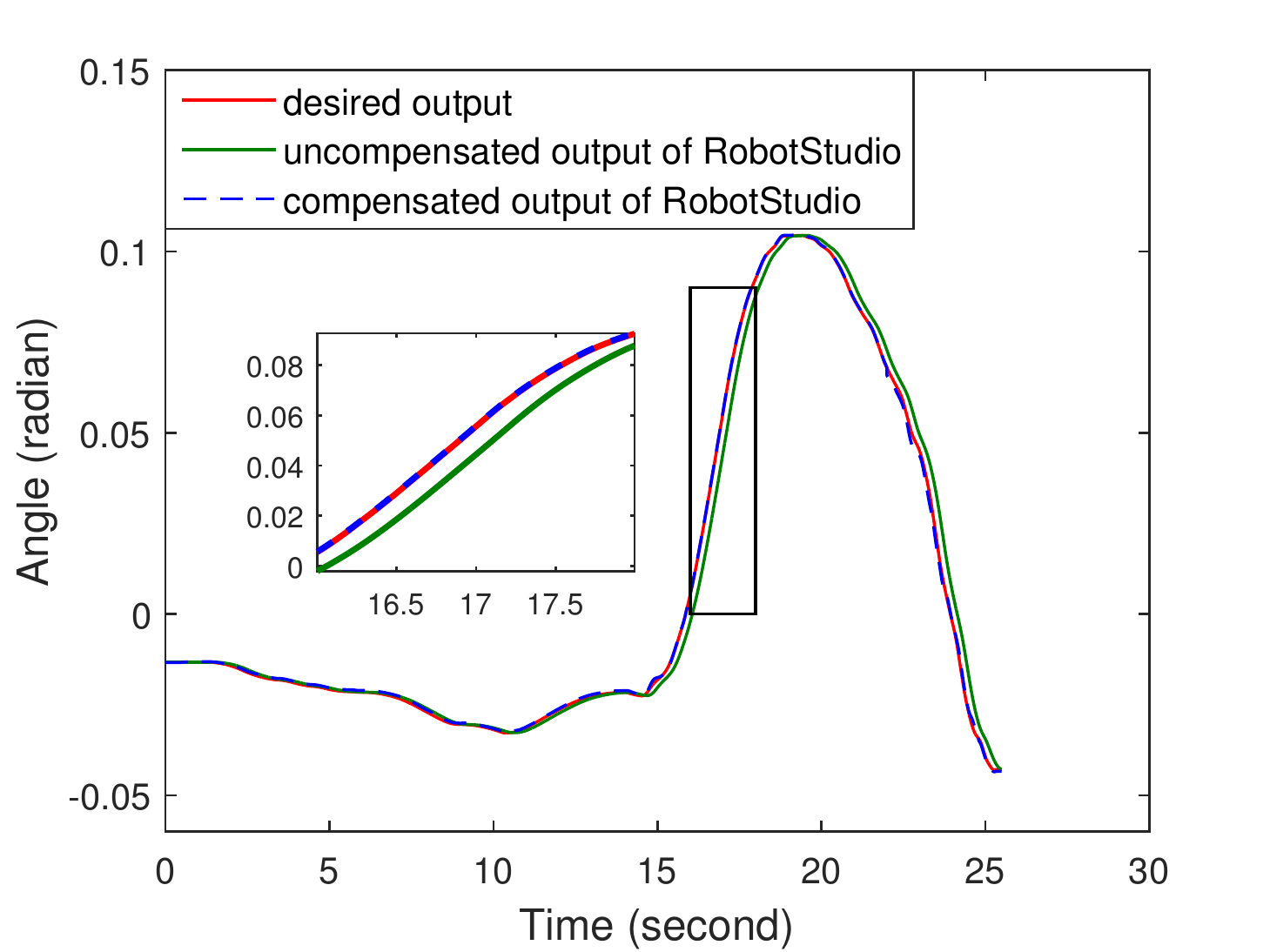}
        \caption{Joint 4}
        \label{fig:J4_RS_moveit_original_final}
    \end{subfigure}    
    \begin{subfigure}[b]{0.32\textwidth}
        \includegraphics[width=\textwidth]{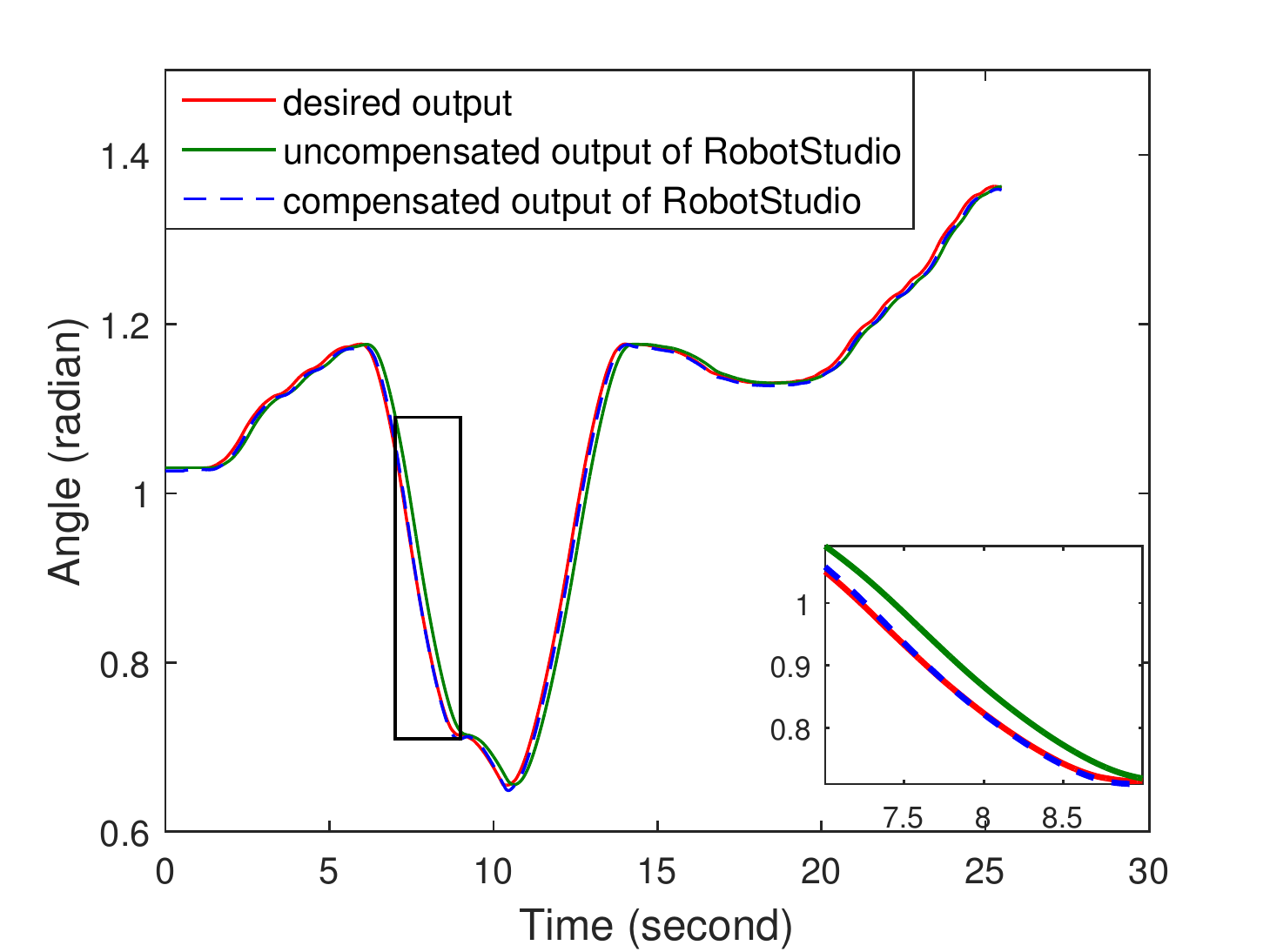}
        \caption{Joint 5}
        \label{fig:J5_RS_moveit_original_final}
    \end{subfigure}    
    \begin{subfigure}[b]{0.32\textwidth}
        \includegraphics[width=\textwidth]{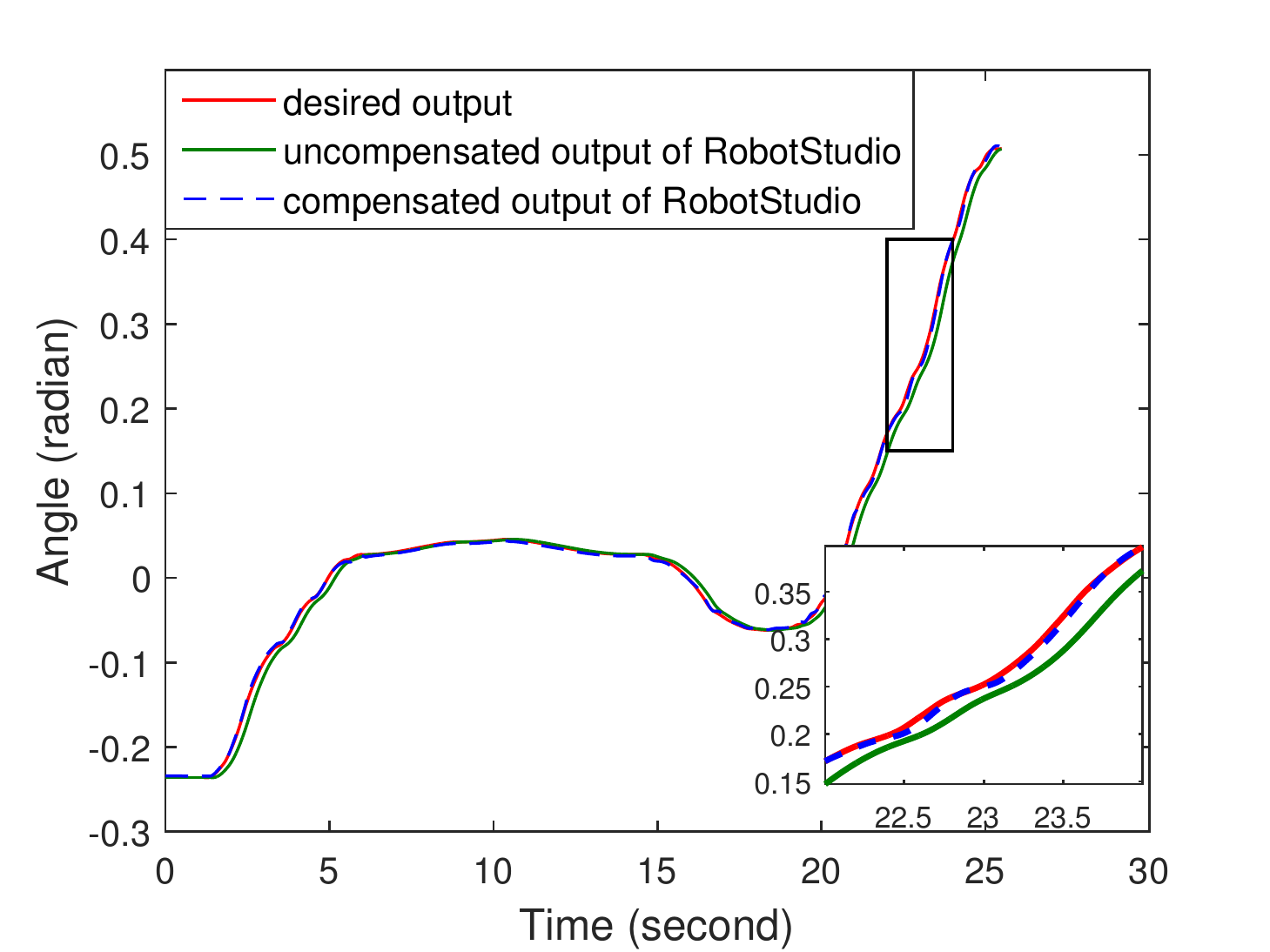}
        \caption{Joint 6}
        \label{fig:J6_RS_moveit_original_final}
    \end{subfigure}
   \caption{Comparison of tracking performance with and without the NNs compensation of the MoveIt! planned joint trajectories in RobotStudio. The inset figure of the zoomed portion demonstrates the improvement of tracking performance.}
   \label{fig:RS_moveit_comparision}
   \vspace{-5pt}
\end{figure*}

To further test the generalization capability of the trained NNs, we conduct another test of tracking 6-dimensional joint trajectories planned by MoveIt! for large structures assembly task~\cite{Chen2018} in RobotStudio. 
The comparison of tracking performance without and with NNs compensation is shown in Fig.~\ref{fig:RS_moveit_comparision}. Table~\ref{moveit_error_joint_path} shows the corresponding tracking errors.

\subsubsection{Cartesian Motion Tracking}

\begin{figure}[tb]
\centering
\includegraphics[width=0.5\textwidth]{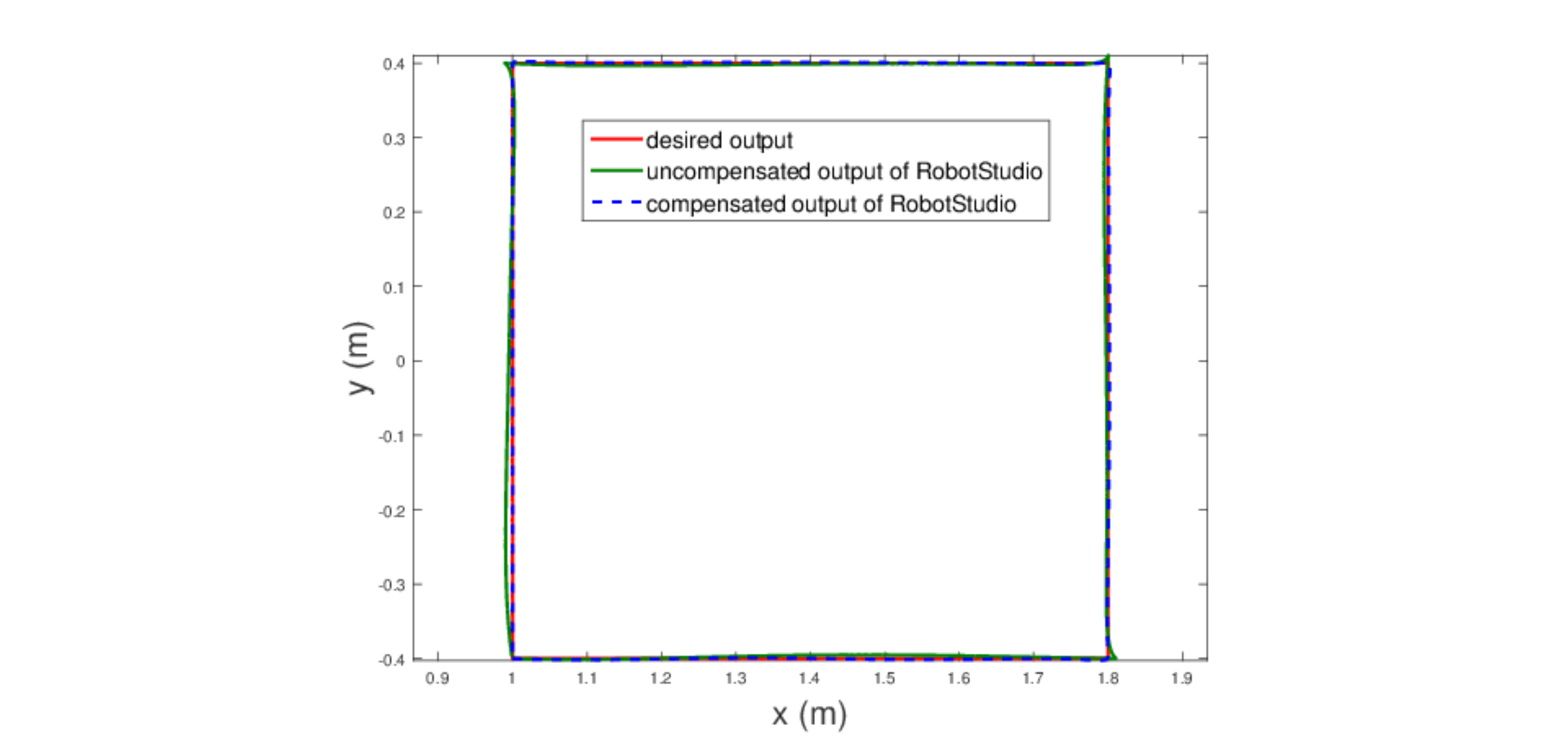}
\caption{Comparison of tracking performance without and with the NN compensation of a Cartesian square trajectory in $x$-$y$ plane with $z$ constant in RobotStudio.}
\label{fig:tracking_sqaure_traj} 
\end{figure}

\begin{figure}[tb]
   \centering
   \begin{subfigure}[b]{0.5\textwidth}
        \includegraphics[width=\textwidth]{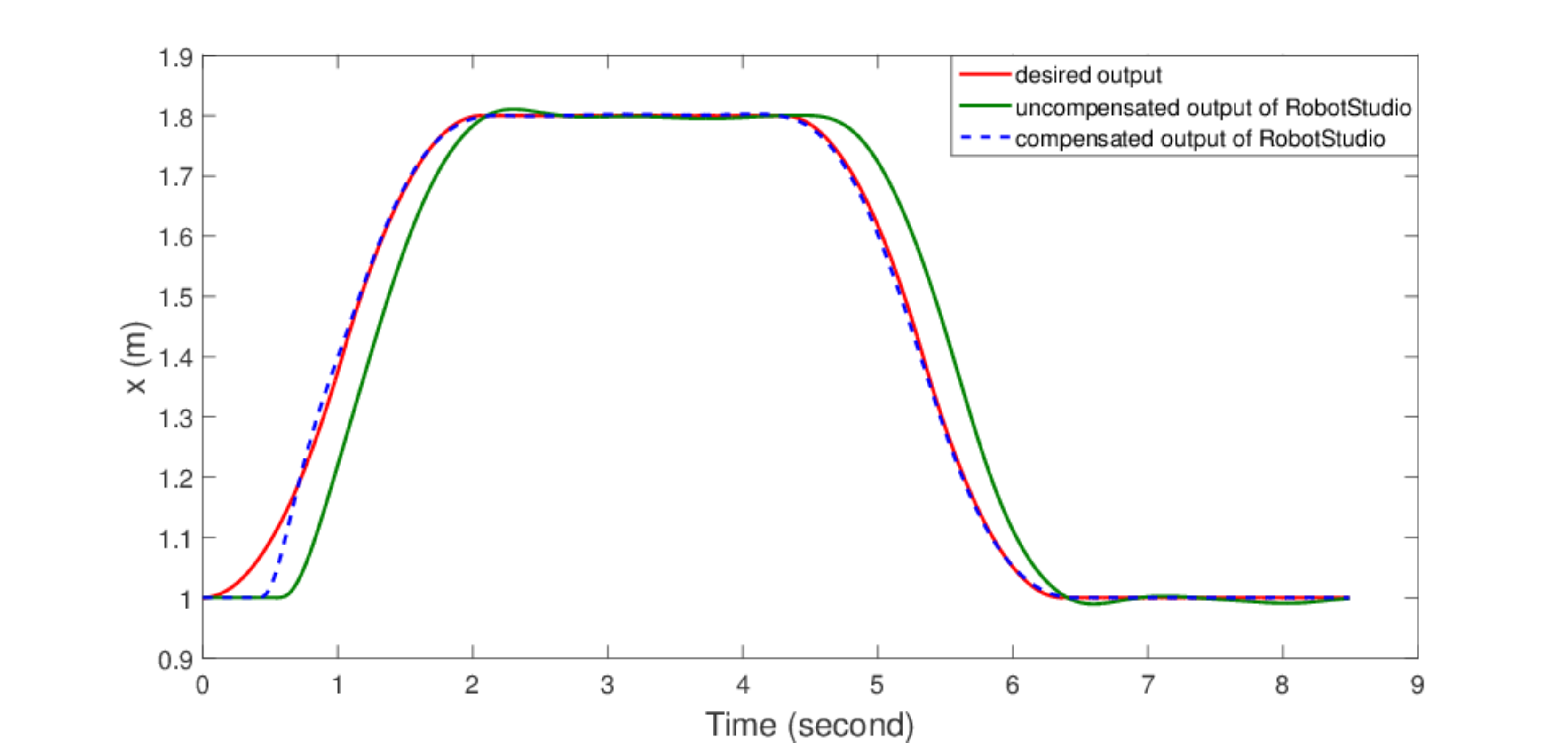}
        \caption{Tracking performance in $x$}
        \label{fig:tracking_x}
    \end{subfigure}
    \hfill
    \begin{subfigure}[b]{0.5\textwidth}
        \includegraphics[width=\textwidth]{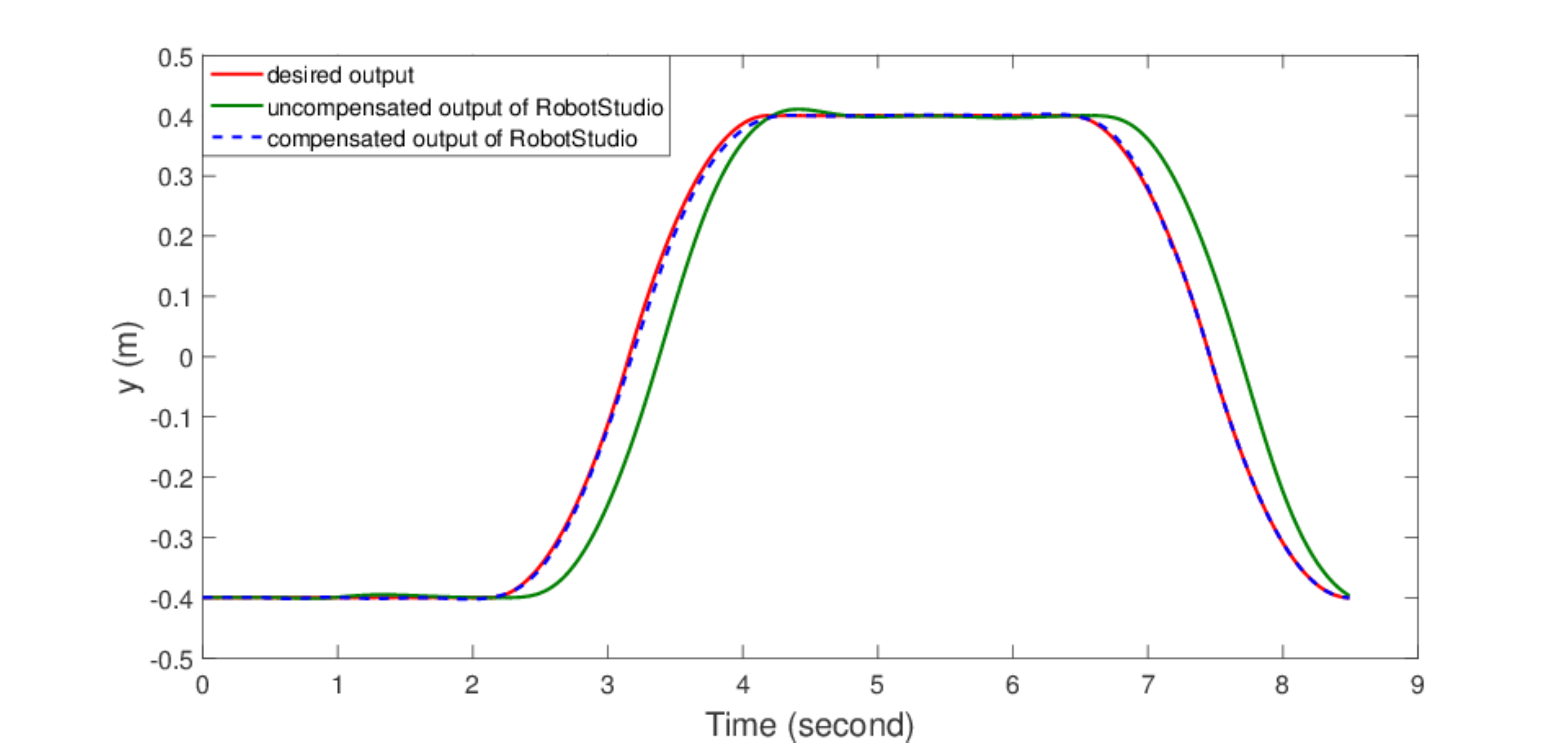}
        \caption{Tracking performance in $y$}
        \label{fig:tracking_y}
    \end{subfigure}
    \hfill
    \begin{subfigure}[b]{0.5\textwidth}
        \includegraphics[width=\textwidth]{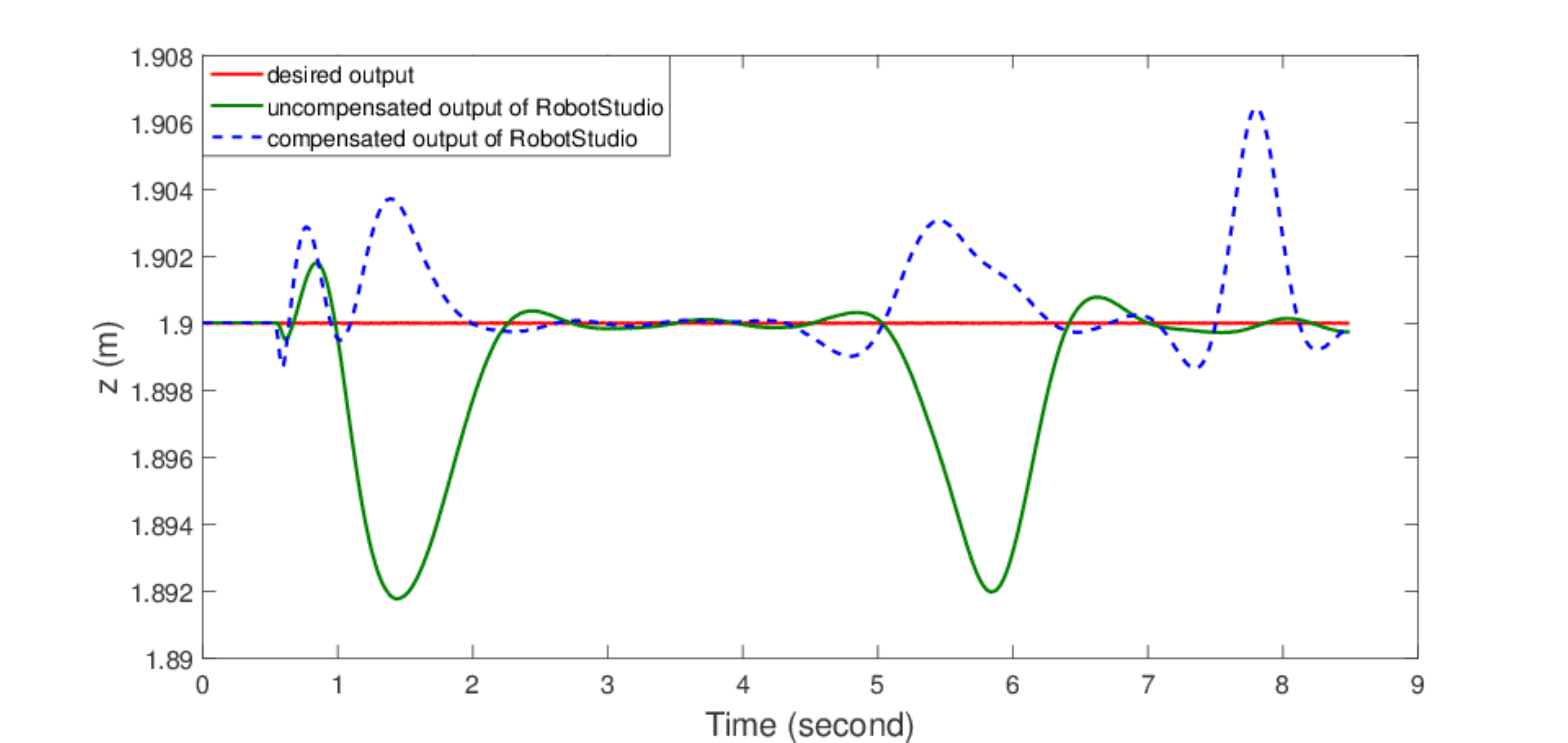}
        \caption{Tracking performance in $z$}
        \label{fig:tracking_z}
    \end{subfigure}
   \caption{The $x, y, z$ tracking performance of Figure~\ref{fig:tracking_sqaure_traj}. With the NN compensation, the tracking errors are significantly reduced in all axes. The $\ell_2$ norm of tracking errors in $x, y, z$ directions reduce from 3.2794~m to 0.7938~m, 3.1475~m to 0.2490~m, and 0.1338~m to 0.0737~m, respectively.}
   \label{fig:tracking_sqaure_traj_subcomponents}
   \vspace{-5pt}
\end{figure}

We also evaluate the performance of the trained NNs on a Cartesian square trajecory (on $x$-$y$ plane, with $z$ constant of 1.9~m) with bounded velocity and acceleration (maximum velocity of 1~m/s and a maximum acceleration of 0.75~$m/s^2$). During the motion, the orientation of the robot end-effector remains fixed. Fig.~\ref{fig:tracking_sqaure_traj} presents that, the trained NNs effectively correct the tracking errors and Fig.~\ref{fig:tracking_sqaure_traj_subcomponents} highlights the tracking performance in three Cartesian directions.

From the figures and tables, we can see that the trained NNs can generalize well to a reasonable path for tracking accuracy improvement in the high fidelity simulator.


\begin{table}[tb]
\caption{$\ell_2$ and $\ell_\infty$ norm of tracking errors of 6-dimensional sinusoidal joint trajectories without and with NNs compensation in RobotStudio.}
\label{error_6_joint_path}
\begin{center}
\begin{tabular}{|c|c|c|}
\hline
Joint & Original Tracking Error (rad) & Final Tracking Error (rad) \\
 & $\ell_2 \hspace{1.5cm} \ell_\infty$ & $\ell_2 \hspace{1.5cm} \ell_\infty$  \\
\hline
1 & 1.4944 \hspace{1cm} 0.0391 & 0.2056 \hspace{1cm} 0.0035 \\
2 & 2.9935 \hspace{1cm} 0.0780 & 0.4993 \hspace{1cm} 0.0065 \\
3 & 6.1128 \hspace{1cm} 0.1563 & 0.8384 \hspace{1cm} 0.0101 \\
4 & 1.5143 \hspace{1cm} 0.0390 & 0.1421 \hspace{1cm} 0.0038 \\
5 & 3.0542 \hspace{1cm} 0.0781 & 0.4555 \hspace{1cm} 0.0065 \\
6 & 5.9629 \hspace{1cm} 0.1561 & 0.8245 \hspace{1cm} 0.0102 \\
\hline
\end{tabular}
\end{center}
\end{table}


\begin{table}[tb]
\caption{$\ell_2$ and $\ell_\infty$ norm of tracking errors of MoveIt! planned joint trajectories without and with NNs compensation in RobotStudio.}
\label{moveit_error_joint_path}
\begin{center}
\begin{tabular}{|c|c|c|}
\hline
Joint & Original Tracking Error (rad) & Final Tracking Error (rad) \\
 & $\ell_2 \hspace{1.5cm} \ell_\infty$ & $\ell_2 \hspace{1.5cm} \ell_\infty$  \\
\hline
1 & 3.5937 \hspace{1cm} 0.1497 & 0.5959 \hspace{1cm} 0.0278 \\
2 & 1.3251 \hspace{1cm} 0.0443 & 0.3992 \hspace{1cm} 0.0180 \\
3 & 1.4463 \hspace{1cm} 0.0493 & 0.2846 \hspace{1cm} 0.0112 \\
4 & 0.3303 \hspace{1cm} 0.0127 & 0.0542 \hspace{1cm} 0.0033 \\
5 & 1.4650 \hspace{1cm} 0.0511 & 0.4773 \hspace{1cm} 0.0154 \\
6 & 0.9629 \hspace{1cm} 0.0392 & 0.1894 \hspace{1cm} 0.0078 \\
\hline
\end{tabular}
\end{center}
\end{table}

\subsection{Experimental Results on IRB6640-180}

\begin{figure}[tb]
   \centering
   \begin{subfigure}[b]{0.49\textwidth}
        \includegraphics[width=\textwidth]{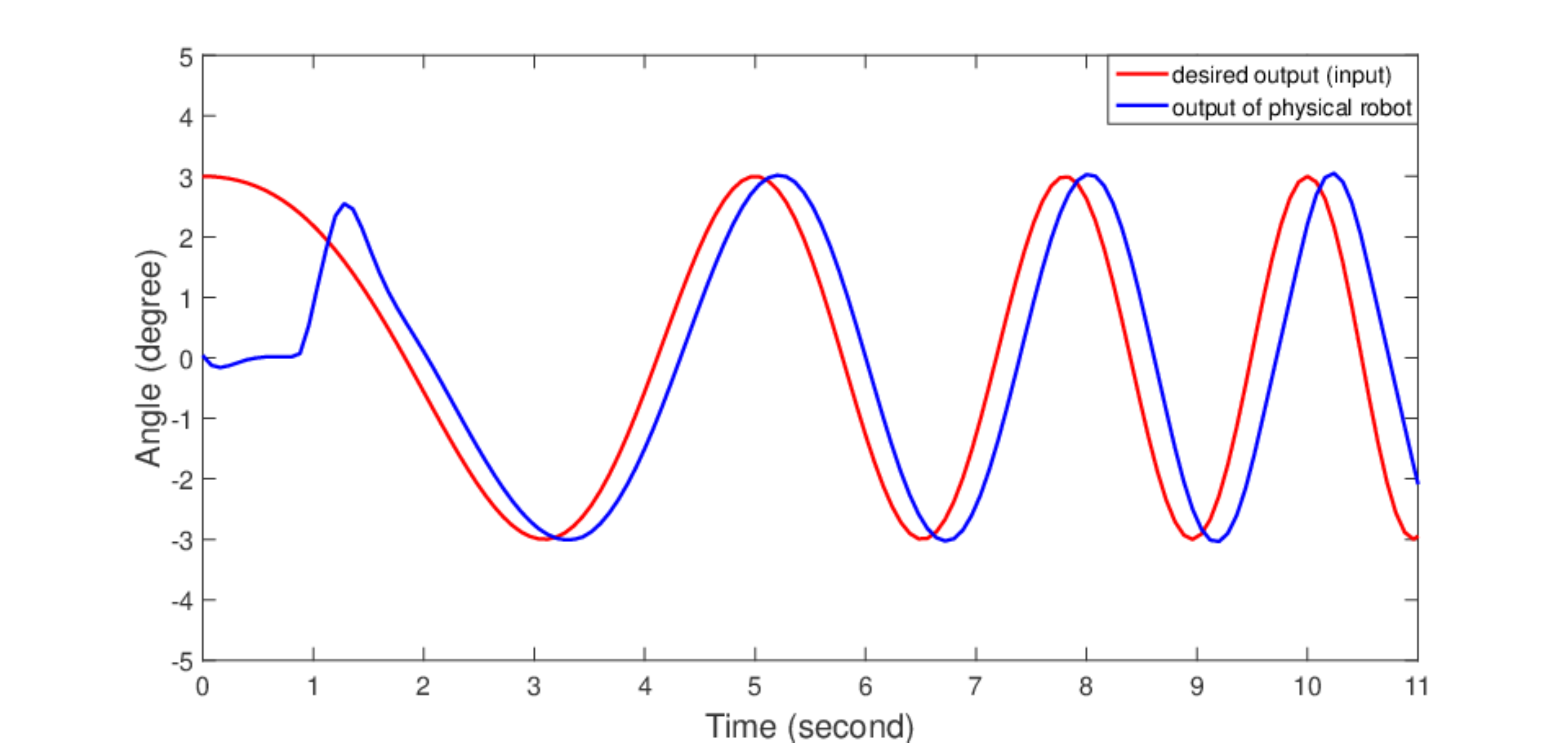}
        \caption{Uncompensated case: desired output (also the input into the inner loop) and the physical robot output.}
        \label{fig:chirp_before_physical}
    \end{subfigure}
    \hfill
    \begin{subfigure}[b]{0.5\textwidth}
        \includegraphics[width=\textwidth]{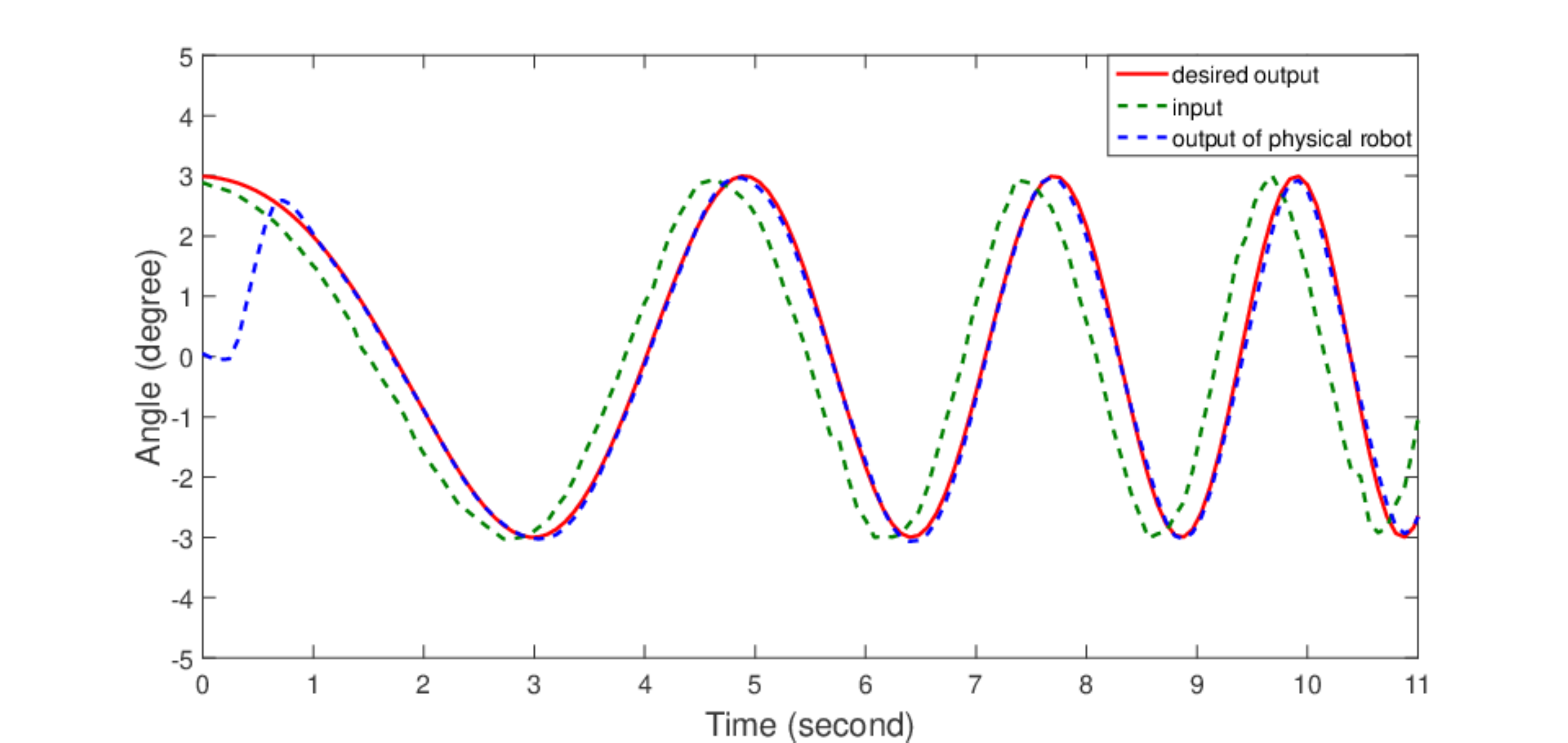}
        \caption{Compensation with  NN:  desired output,  physical robot output, and the input generated by the NN.}
        \label{fig:chirp_final_physical}
    \end{subfigure}
   \caption{Comparison of tracking performance of the same chirp signal as the one in simulation without and with the NN compensation for the joint 1 of the physical robot.}
   \label{fig:chirp_physical}
   \vspace{-5pt}
\end{figure}

 
We tested the tracking performance of the same chirp 
signal as we did in simulation for joint 1 of the physical robot, as shown in Fig.~\ref{fig:chirp_physical}. 
The command input is filtered by the NNs obtained by the transfer learning. Fig.~\ref{fig:chirp_before_physical} 
shows the uncompensated output with the robot output lagging behind the desired output. Fig.~\ref{fig:chirp_final_physical} 
shows that, by compensating for the lag effect, the command input generated by the NNs can drive the output to the desired output closely after a brief initial transient. We found that for trajectories with low velocity profiles (like the chirp signal and the sinusoid in Fig.~\ref{fig:sin_mag2_omega2}), there exists negligible difference between the filtered inputs of NNs trained by simulation data and NNs obtained by transfer learning. 

\begin{figure}[tb]
   \centering
   \begin{subfigure}[b]{0.5\textwidth}
        \includegraphics[width=\textwidth]{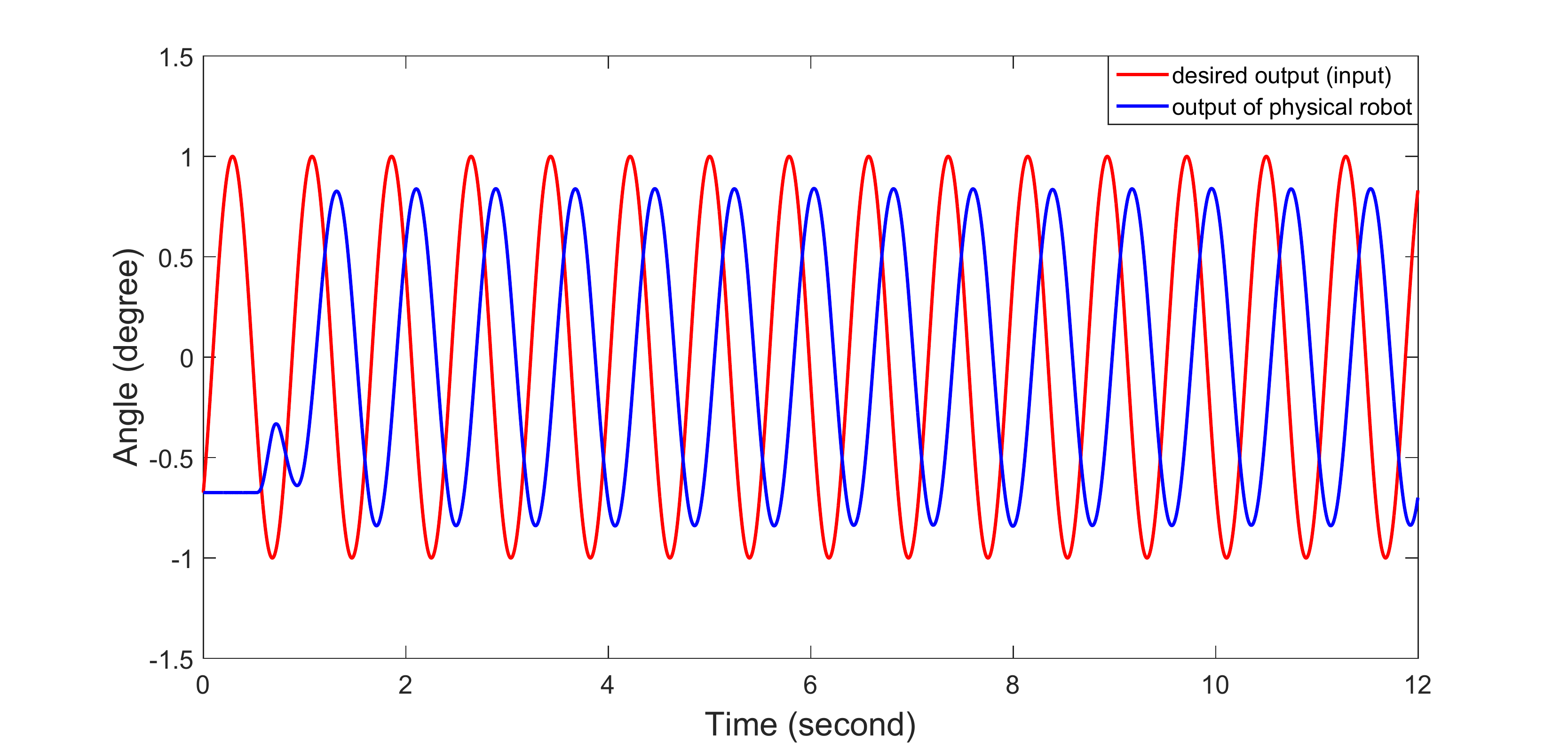}
        \caption{Uncompensated case}
        \label{fig:PR_sin_mag1_omega8}
    \end{subfigure}
    \hfill
    \begin{subfigure}[b]{0.5\textwidth}
        \includegraphics[width=\textwidth]{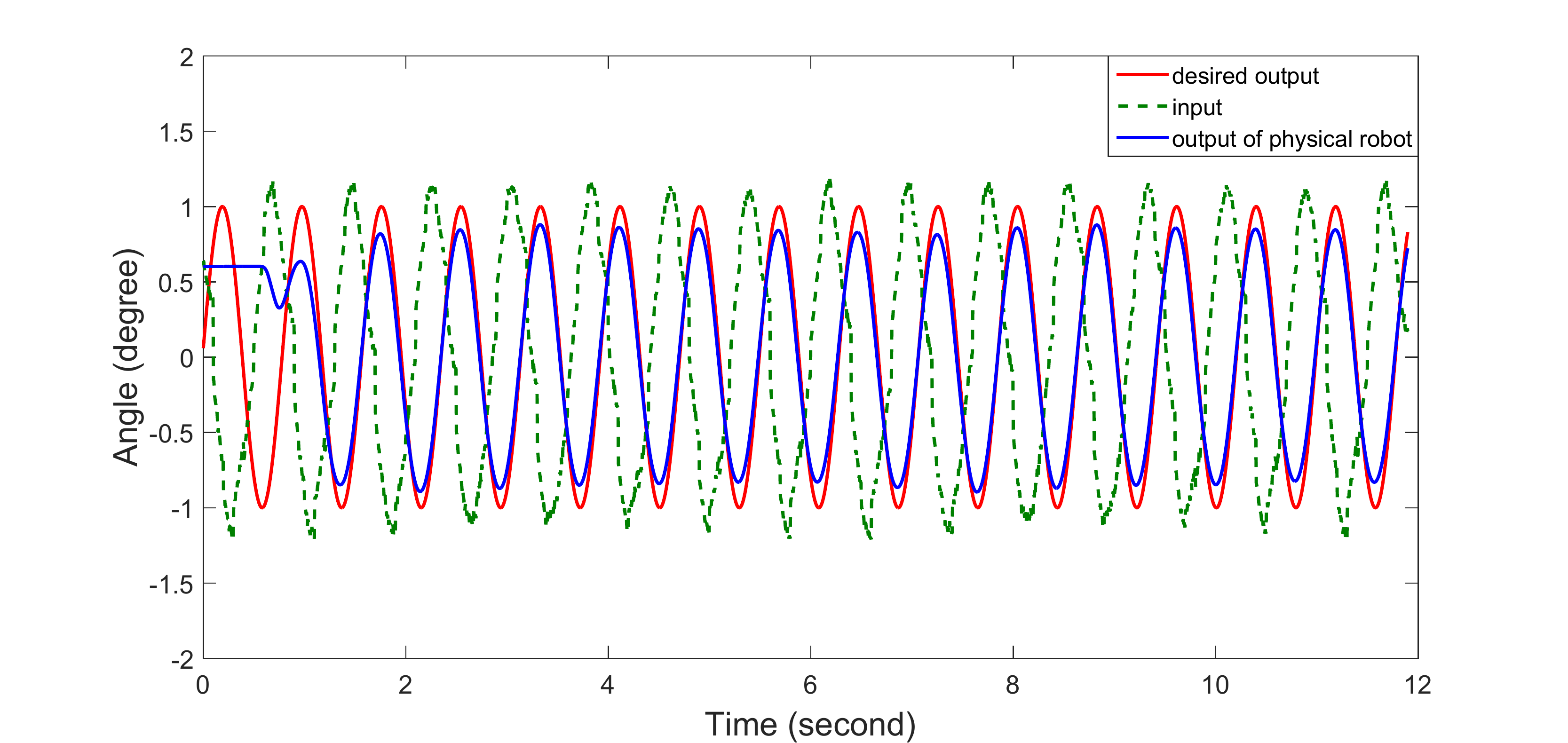}
        \caption{Compensation using NN trained by simulation data}
        \label{fig:PR_sin_mag1_omega8_final_simulatedNN}
    \end{subfigure}
    \hfill
    \begin{subfigure}[b]{0.5\textwidth}
        \includegraphics[width=\textwidth]{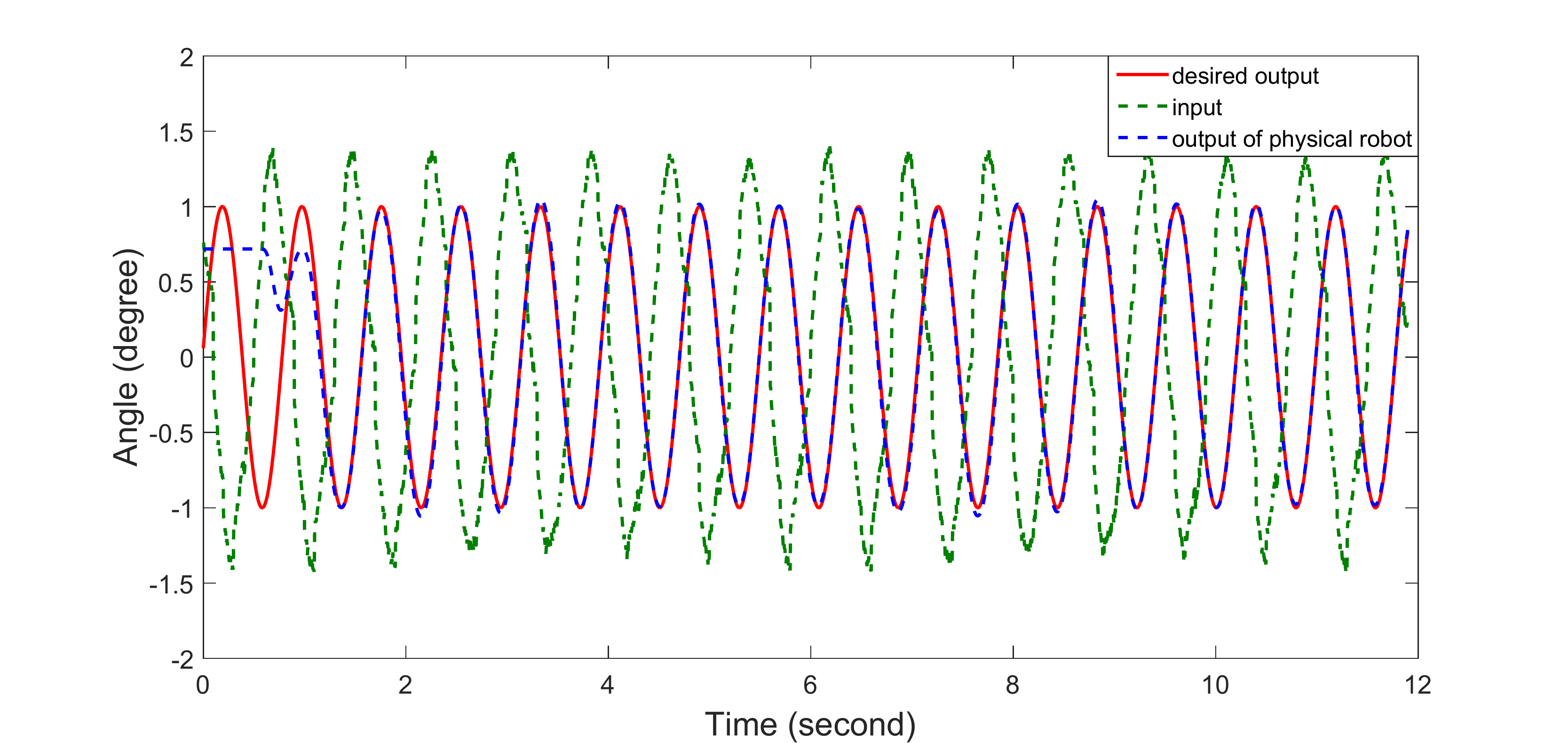}
        \caption{Compensation using NN after transfer learning}
        \label{fig:PR_sin_mag1_omega8_final_transferredNN}
    \end{subfigure}
   \caption{Comparison of tracking performance of a sinusoidal path without NN compensation, with compensation by the NN trained by the simulation data and NN obtained by transfer learning, for joint 1 of the physical robot. The $\ell_2$ norm of tracking error of each case is 58.6822$^{\circ}$, 14.0066$^{\circ}$ and 13.6826$^{\circ}$, respectively.   }
   \label{fig:sin_mag1_omega8_PR}
   \vspace{-5pt}
\end{figure}

However, for trajectories with large velocity and acceleration profiles, transfer learning plays a key role in generating an optimal input to reduce trajectory tracking error for the physical robot. Fig.~\ref{fig:sin_mag1_omega8_PR} illustrates the tracking of a sinusoidal signal with angular frequency of 8 rad/s (with different amplitude and phase from training data). From the comparison between Figs.~\ref{fig:PR_sin_mag1_omega8_final_simulatedNN} and~\ref{fig:PR_sin_mag1_omega8_final_transferredNN}, we can see that transfer learning is necessary to correctly compensate for the phase lagging and magnitude degradation for the physical robot.





\subsection{RobotStudio Simulation Results on Other Robots}

\begin{figure}[tb]
   \centering
   \begin{subfigure}[b]{0.5\linewidth}
        \includegraphics[width=\linewidth]{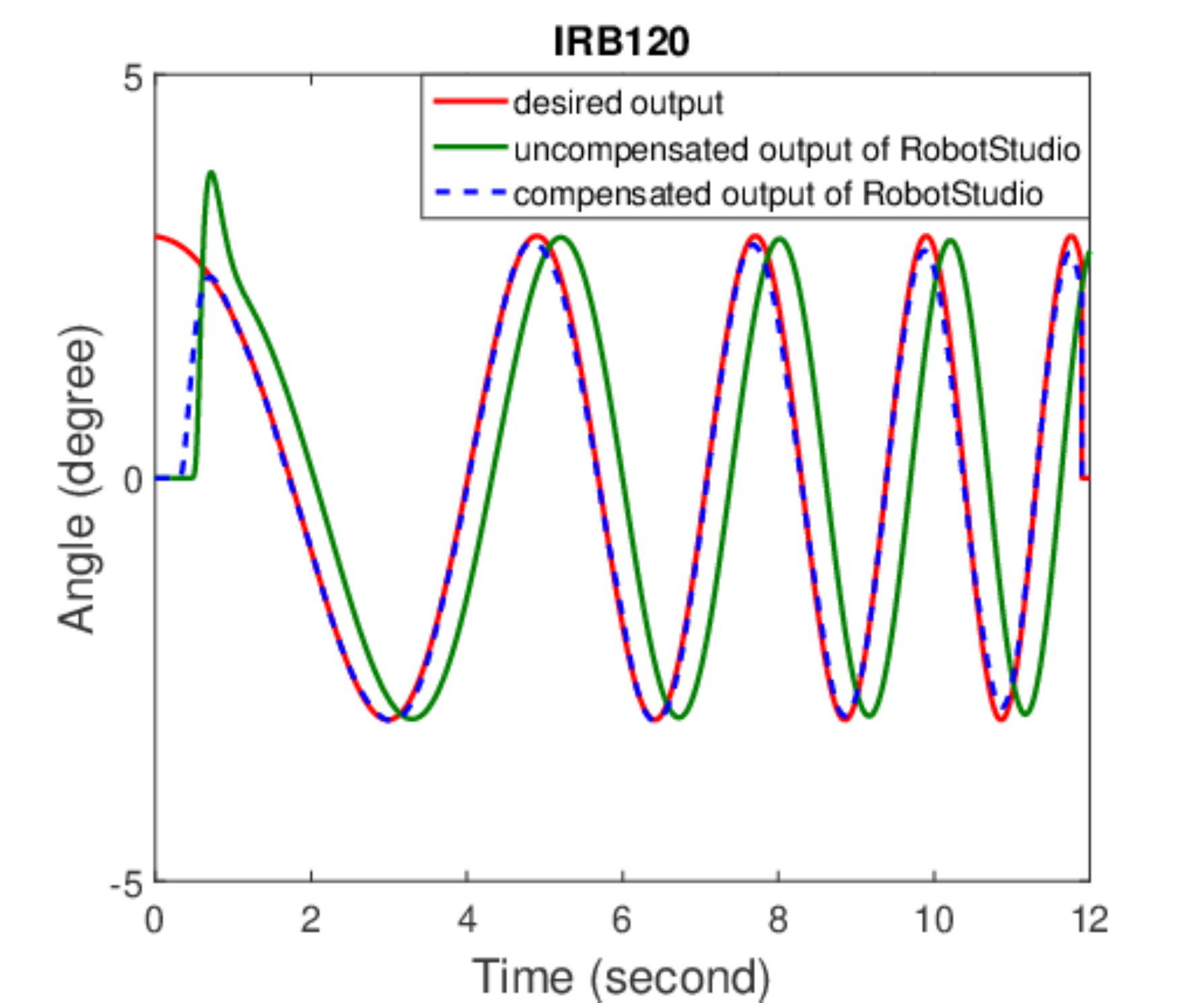}
        \caption{IRB120 tracking a chirp}
        \label{fig:chirp_irb120_final_sqaure}
    \end{subfigure}
    \hspace{-0.3cm}
    \begin{subfigure}[b]{0.5\linewidth}
        \includegraphics[width=\linewidth]{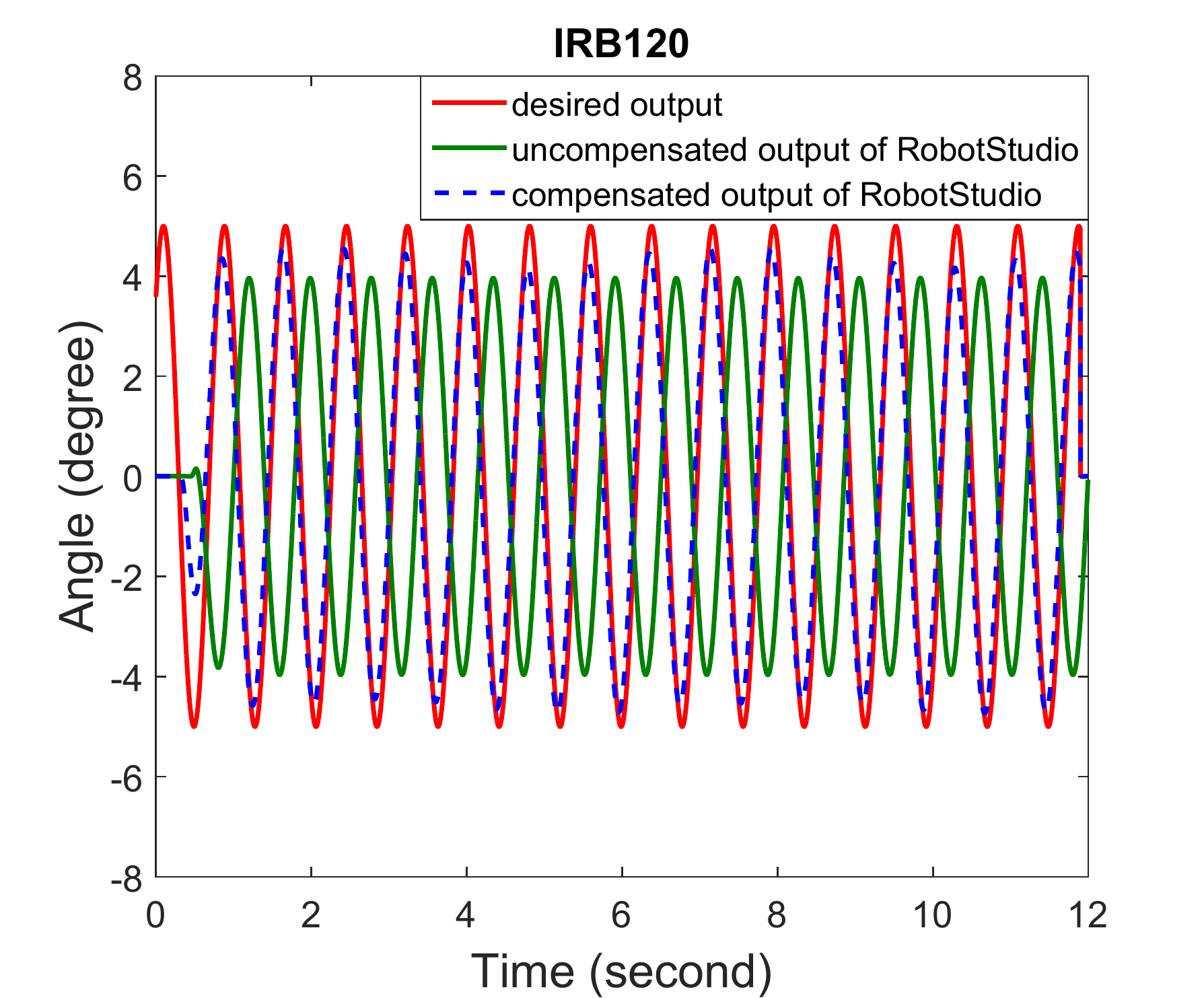}
        \caption{IRB120 tracking a sinusoid}
        \label{fig:sin_omega8_magnitude5_irb120_final_square}
    \end{subfigure}    
  
    \begin{subfigure}[b]{0.5\linewidth}
        \includegraphics[width=\linewidth]{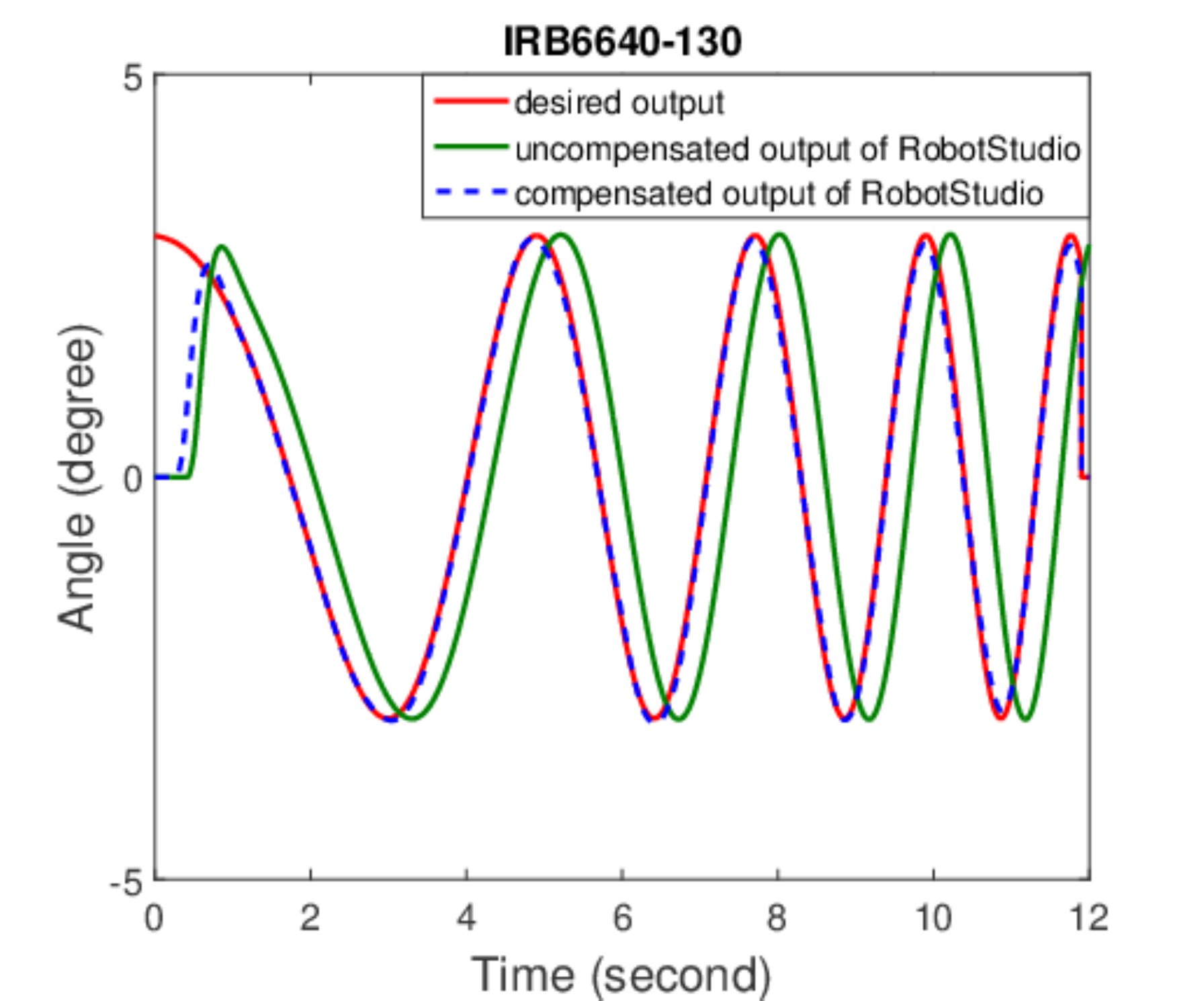}
        \caption{IRB6640-130 tracking \\ a chirp}
        \label{fig:chirp_irb6640_130_final_square}
    \end{subfigure}    
    \hspace{-0.3cm}
    \begin{subfigure}[b]{0.5\linewidth}
        \includegraphics[width=\linewidth]{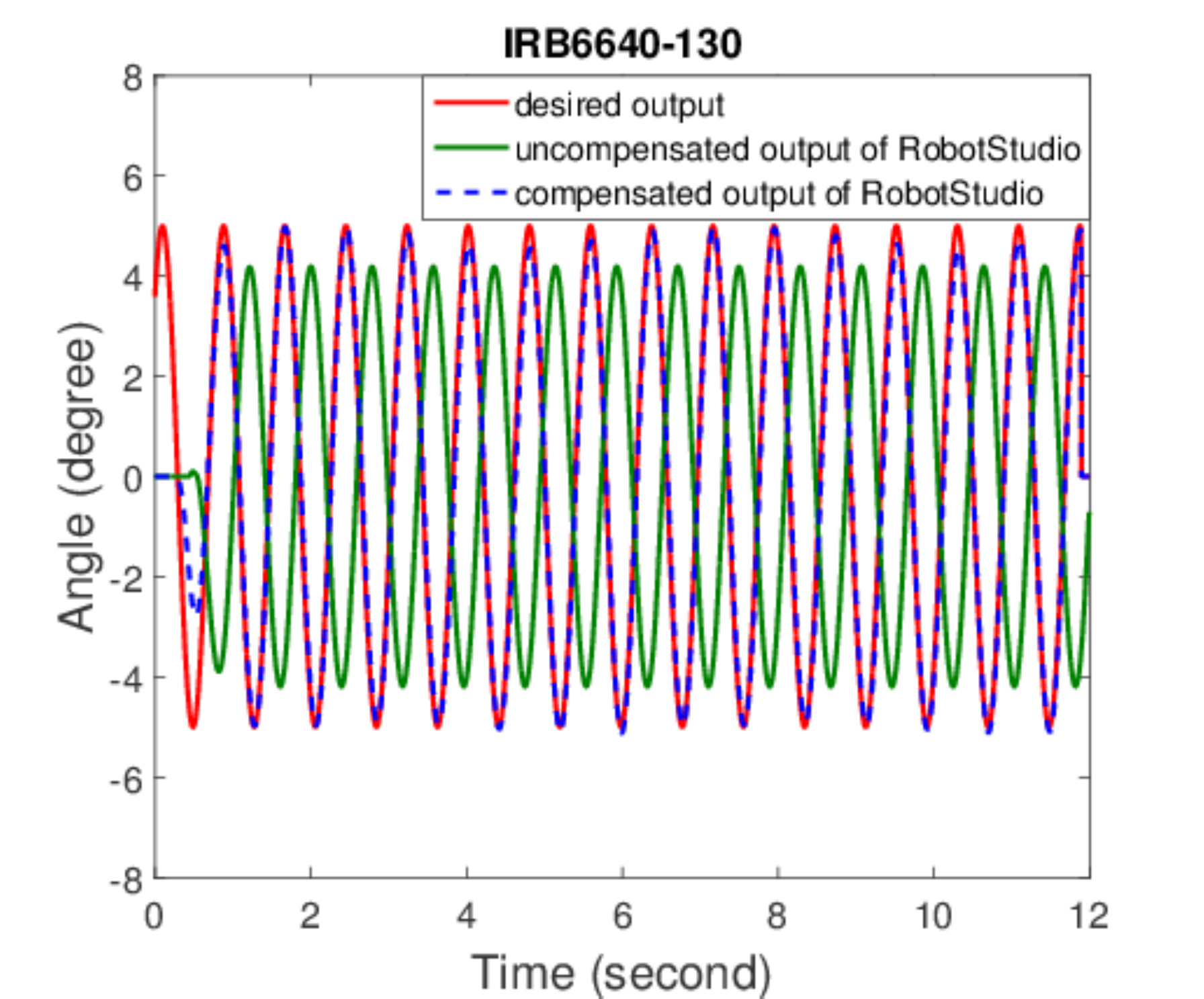}
        \caption{IRB6640-130 tracking a \\ sinusoid}
        \label{fig:sin_omega8_magnitude5_irb6640_130_final_square}
    \end{subfigure}    
   \caption{Comparison of tracking performance with and without the NNs compensation of a chirp and a sinusoidal joint trajectories in RobotStudio on IRB120 and IRB6640-130.}
   \label{fig:test_on_two_additional_robots}
   \vspace{-5pt}
\end{figure}

One natural question that will come up is whether the trained NNs can work on different robot models. We applied the original trained NNs by simulation to two robot models IRB120 and IRB6640-130, and tested their tracking on a chirp trajectory (the same one as we use in Section~\ref{Single Joint Motion Tracking}) and a sinusoidal signal with angular frequency of 8~rad/s and magnitude of 5 degrees in RobotStudio, as shown in Fig.~\ref{fig:test_on_two_additional_robots}.
  
From the figure, we can see that the trained NNs can effectively compensate for the lag effect and amplitude discrepancies for these two robot models. For tracking of the chirp signal, NNs compensate for the delay well for these robots and a
possible explanation for this  could be that EGM has the same time delay on all robot models. For tracking of the sinusoid signal with obvious nonlinear effects, NNs compensate for both the delay and the magnitude degradation well, since their inner loop dynamics has similar behavior to IRB6640-180 robot.

\subsection{Limitations}
\begin{figure}[tb]
\centering
\setlength{\unitlength}{0.012500in}%
\includegraphics[width=0.5\textwidth]{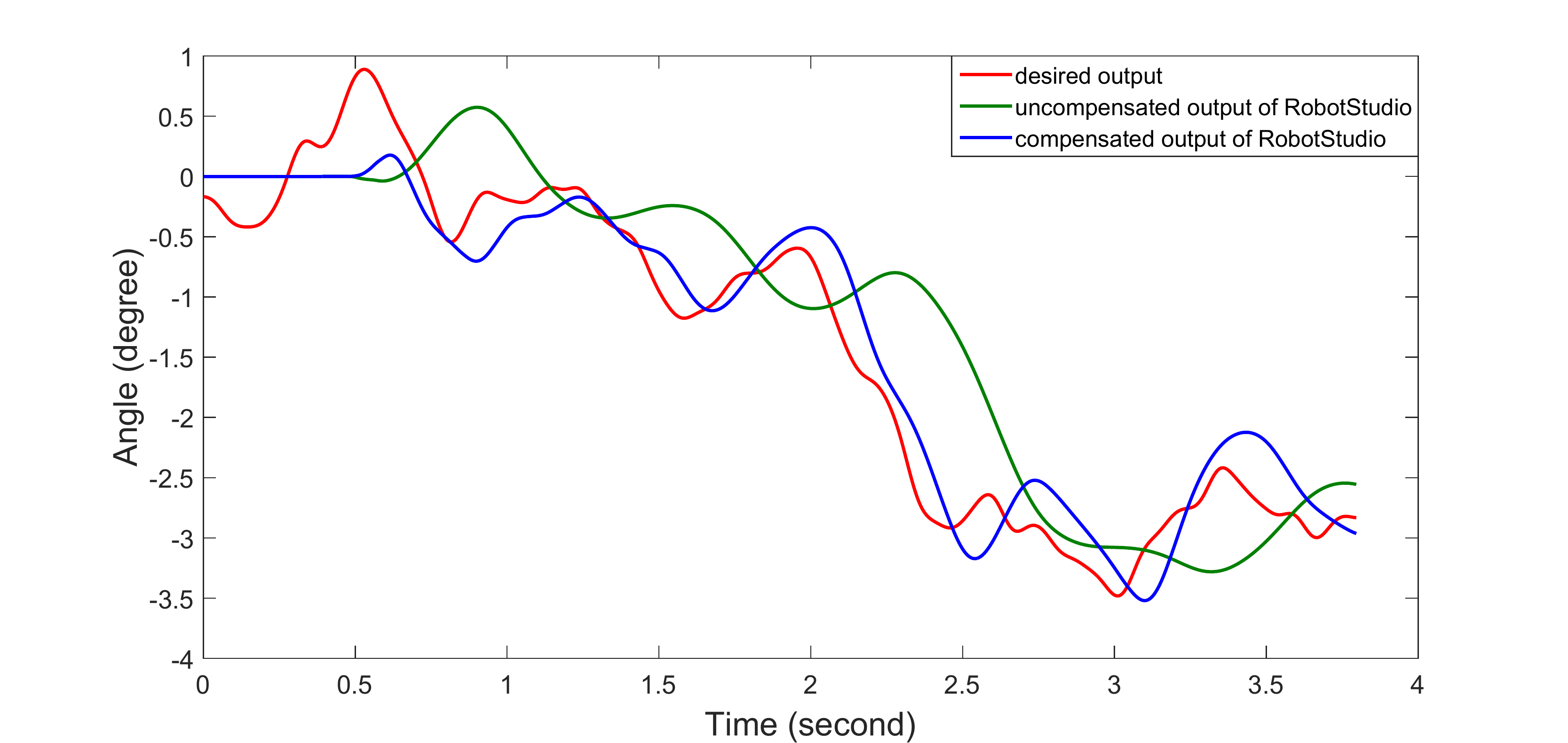}
\caption{Comparison of tracking performance with and without the NNs compensation of a random joint trajectory in RobotStudio.}
\label{fig:random_path} 
\end{figure}

Firstly, our approach assumes the robot internal dynamics and delays are deterministic and repeatable, and the environment is still but for some robots with flexible joints and working outdoors like Baxter and quadrotors, the assumption will not hold anymore. To address this problem and improve the robustness of the controller, we may combine $L_1$ adaptive control and ILC in response of disturbances and uncertainties. Secondly, we have a high-fidelity simulator which facilitates the ILC implementation and data collection process. But for robots that does not have a high-fidelity dynamic simulator such as Baxter and Motoman, we may use a two-NN architecture that one NN emulates a simulator and produces the dynamic response and another NN approximates the inverse dynamics. Finally, since the training data of the NNs consist of smooth trajectories with bounded velocity and acceleration such as sinusoids, the NNs may not work as well on random trajectories with lots of noise. Fig.~\ref{fig:random_path} 
shows the tracking result of a random trajectory for joint 1 of the simulated robot. From the figure, we can see that the trained NNs still work for compensating for the lag effect and amplitude degradation, but not as well as for smooth trajectories as shown in previous subsections.

\section{Conclusions and Future Work}
\label{conclusions and future work}
In this paper, we propose the approach of combining multi-layer NNs and ILC to achieve high-performance tracking of a 6-dof industrial robot. Large amount of data for NNs training are collected by ILC for multiple trajectories through a high-fidelity physical simulator. After training, the NNs can generalize well to other desired trajectories and improve tracking performance significantly without on-line iterations. We use transfer learning to narrow the reality gap, and further explore the generality of the NNs on two different robot models. The feasibility of our approach is demonstrated by simulation and physical experiments.

Future work includes the development of predictive motion and force controllers based on the trained NNs. 
Moreover, we can add feedback components into the control design to further improve the tracking accuracy and robustness of the controller. Finally, it is worthwhile to verify the possibility of the NNs to be served as a general compensator.

\bibliographystyle{IEEEtran}
\bibliography{main}

\end{document}

%% file: wenmacro.tex
\def\beq{\begin{equation}}
\def\eeq{\end{equation}}
\def\beqa{\begin{eqnarray}}
\def\eeqa{\end{eqnarray}}
\def\ba{\begin{eqnarray*}}
\def\ea{\end{eqnarray*}}
\def\bc{\begin{center}}
\def\ec{\end{center}}
\def\ma{\left[ \begin{array}}
\def\ema{\end{array}\right]}

\def\tr{ ^T }

\def\inv{ ^{-1} }

\def\norm#1{\left\| {#1} \right\|}

\def\rr#1{\mathbb{R}^{#1}}

\def\ybar{{\underline y}}
\def\ydbar{{\underline y}_d}
\def\qdbar{{\underline q}_d}
\def\eybar{{\underline e}_y}
\def\ubar{{\underline u}}

\def\sup#1{^{({#1})}}

%% file: robmacro.tex
\def\rr#1{\mathbb{R}^{#1}}

\def\calG{{\cal G}}